\documentclass{optica-article}

\journal{opticajournal} 

\articletype{Research Article}

\usepackage{graphicx}
\usepackage{amsmath}
\usepackage{hyperref}
\usepackage{xcolor}
\usepackage{subcaption}
\usepackage{booktabs}
\usepackage{tikz}
\usetikzlibrary{arrows.meta,positioning,fit,calc,shapes.misc}

\begin{document}

\title{Vision Mamba for Permeability Prediction of Porous Media}

\author{Ali Kashefi\authormark{1,$\dag$}, Tapan Mukerji\authormark{2}}

\address{Stanford University, Stanford, CA 94305, USA\\
\email{\authormark{1}kashefi@stanford.edu}
\email{\authormark{2}mukerji@stanford.edu}
\authormark{$\dag$}The corresponding author}

\begin{abstract*} 
Vision Mamba has recently received attention as an alternative to Vision Transformers (ViTs) for image classification. The network size of Vision Mamba scales linearly with input image resolution, whereas ViTs scale quadratically, a feature that improves computational and memory efficiency. Moreover, Vision Mamba requires a significantly smaller number of trainable parameters than traditional convolutional neural networks (CNNs), and thus, they can be more memory efficient. Because of these features, we introduce, for the first time, a neural network that uses Vision Mamba as its backbone for predicting the permeability of three-dimensional porous media. We compare the performance of Vision Mamba with ViT and CNN models across multiple aspects of permeability prediction and perform an ablation study to assess the effects of its components on accuracy. We demonstrate in practice the aforementioned advantages of Vision Mamba over ViTs and CNNs in the permeability prediction of three-dimensional porous media. We make the source code publicly available to facilitate reproducibility and to enable other researchers to build on and extend this work. We believe the proposed framework has the potential to be integrated into large vision models in which Vision Mamba is used instead of ViTs.
\end{abstract*}

\begin{backmatter}
\bmsection{Keywords}
Vision Mamba; Vision transformer; Convolutional neural network; Porous media; Permeability
\end{backmatter}

\section{Introduction and motivation} 
\label{Sect1}

Porous media play a central role across diverse scientific and industrial domains, including digital rock physics \cite{ANDRApart1,ANDRApart2,zhu2025joint}, membrane systems \cite{LIANG2023116359,ferro2025numerical}, geological carbon storage \cite{BLUNT2013197,yang2025novel}, and medicine \cite{PorousMedicine,Das2018,farooq2025boundary}. Conventional investigations use numerical simulations and laboratory experiments to analyze porous media and to obtain their physical and geometric characteristics. Although both approaches are valuable, they are resource-intensive, requiring substantial computation, specialized lab instrumentation, and considerable wall-clock time. To reduce this burden, deep learning within the broader machine-learning paradigm can accelerate tasks such as segmentation of porous media \cite{bihani2022mudrocknet,lee2023deep,han2024advanced,wang2025reunet} and the prediction of porous-medium properties, including permeability \cite{meng2023transformer,PermCurve,kashefi2021PointNetPorousMedia,Mingliang1,Hong2020RapidEstimation,WU20181215,MASROOR2023,consideringorganicmatter,kashefi2024novelFNO}, porosity \cite{graczyk2020predicting}, elasticity \cite{chung2024prediction,liu2025deepElasticity}, and effective diffusivity \cite{wu2019predicting}. Moreover, deep learning configurations can predict pore-scale fields such as velocity and pressure \cite{santos2020poreflow,kamrava2021simulating,kashefi2023PorousMediaPIPN}. Additionally, generative deep learning models are used for porous-media reconstruction \cite{Liu2022Generative,guan2021reconstructing,phan2024generating,baishnab20253d}. In the present work, we focus on predicting the permeability of porous media from digital rock images using supervised deep-learning frameworks.

From a computer-science perspective, a variety of deep-learning frameworks have long been applied to permeability prediction in porous media, each with its own advantages and limitations. We briefly review these approaches and then explain how our proposed deep-learning framework addresses several of their challenges while introducing new capabilities. Convolutional neural networks (CNNs) and CNN-based variants such as ResNet \cite{he2016residual} have been widely used to predict permeability from 2D and 3D representations of porous media \cite{PermCurve,Mingliang1,Hong2020RapidEstimation,TANG2022127473}. These models often achieve strong accuracy with relatively simple architectures (compared with other models that will be mentioned later in this paragraph). However, they typically require a large number of learnable parameters, often more than the alternatives we will discuss later. They also operate on fixed input resolutions; a network trained on cubes of one size generally expects test data of the same size. Point-cloud neural networks, such as PointNet \cite{qi2017pointnet,KASHEFI2025PointNetGraphicKAN} and PointNet++ \cite{NIPS2017PointNetPlus}, are another family of deep-learning frameworks used for permeability prediction \cite{kashefi2021PointNetPorousMedia}. In this setup, the boundary between pore and grain phases of the porous medium is represented as a point cloud. The main advantage of this approach compared with CNNs is that it dramatically reduces the dataset size, since the full volumetric cubes are no longer needed \cite{kashefi2021PointNetPorousMedia}. Additionally, although models are usually trained with the same number of points per point cloud within a batch (each batch can still contain point clouds with different numbers of points), at test time, the number of points can vary. However, preprocessing is required to convert volumetric images of porous media into point-cloud data. Furthermore, if the number of boundary points varies drastically across the dataset, the training procedure may face additional challenges, both in implementation and in loss-function convergence. Fourier neural operators (FNOs) have also been used for permeability prediction \cite{kashefi2024novelFNO}. FNOs are invariant to input image size \cite{li2020fourierFNO}; leveraging this property, they can be trained on porous media of different sizes simultaneously and have shown strong generalizability to unseen sizes. However, FNOs can be prone to overfitting on the training data. Another limitation is their sensitivity to hyperparameters, especially the number of Fourier modes, which introduces additional challenges for training and fine-tuning \cite{kashefi2024novelFNO}. Vision Transformers (ViTs) \cite{dosovitskiy2021imageworth16x16words}, as another deep-learning architecture, have been applied to predicting the permeability of porous media \cite{geng2024swin,temizel2025permeability,MENG2023104520}. In several settings, ViTs achieve competitive or superior accuracy with comparable, or sometimes fewer, trainable parameters than CNNs \cite{MENG2023104520}. Moreover, unlike standard CNNs, vanilla ViTs with full self-attention can look across the entire image from the very first layer, so they pick up long-range patterns early; CNNs usually need many layers or operations like pooling or dilated convolutions to see that much of the image. 



Mamba \cite{gu2024mambalineartimesequencemodeling} was introduced as an alternative to Transformers \cite{vaswani2023attentionneed}. Building on this line of work, Vision Mamba \cite{zhu2024visionmambaefficientvisual} was also introduced as an alternative to Vision Transformers \cite{dosovitskiy2021imageworth16x16words}. One of the main advantages of Vision Mamba over ViTs is that it scales linearly (rather than quadratically) with the number of tokens. This motivates us to propose a deep learning framework based on Vision Mamba for predicting the permeability of porous media. Vision Mamba has been so far used for several key applications in vision tasks \cite{zhu2024visionmambaefficientvisual,DBLP} such as image detection \cite{liu2024vmambavisualstatespace}, medical image classification \cite{yue2024medmambavisionmambamedical}, remote sensing \cite{bao2025visionmambaremotesensing}, ocean engineering of underwater vehicles \cite{liu2025mambaOcean}, medical image segmentation \cite{wang2025comprehensiveanalysismamba3d}, and medical video segmentation \cite{yang2024vivimvideovisionmamba}. It is important to note that Mamba and its vision counterpart are emerging as alternatives to Transformers and ViTs and are increasingly used as building blocks in large language and large vision models \cite{openai2024gpt4,chang2023surveyevaluationlargelanguage,geminiteam2024gemini15unlockingmultimodal,Kashefi2023ChatGPT,Kashefi2024misleading,basant2025nvidia}. Demonstrating that Vision Mamba can predict properties of three-dimensional porous media is therefore significant, as it indicates a pathway to incorporating this task into future large models that perform multiple functions, including permeability estimation from volumetric data.

The key contributions of this study are as follows.

\begin{itemize}
 
    \item We introduce a novel neural network, based on the Vision Mamba architecture, for predicting the permeability of voxelized porous media.

    \item The proposed network leverages Vision Mamba to achieve linear scaling with token size (or similarly patch size), whereas ViTs scale quadratically.

    \item Leveraging Vision Mamba significantly reduces trainable parameters compared to CNNs, improving memory efficiency.


    \item The code and documentation are released as open-source to support reproducibility, educational purposes, and future extensions of this work.

\end{itemize}

We now outline the structure of the remainder of this research paper. In Sect. \ref{Sect2}, we describe the generation and collection of three-dimensional porous media and the computation of their permeability by numerically solving the Stokes equations for the proposed supervised deep-learning framework. In Sect. \ref{Sect3}, we present the Vision Mamba architecture, adapted to predict a volumetric property, here, the permeability of 3D porous media. Training of Vision Mamba and the hyperparameter settings are explained in Sect. \ref{Sect4}. We then discuss the results in Sect. \ref{Sect5}, including the performance of Vision Mamba, its comparison with CNNs and ViTs, and the ablation studies. Finally, Section \ref{Sect6} provides a summary and potential directions for future research.


\section{Data generation}
\label{Sect2}

\begin{figure*}[htp]
    \centering
    \begin{subfigure}{0.32\textwidth}
        \centering
         \includegraphics[width=\linewidth]{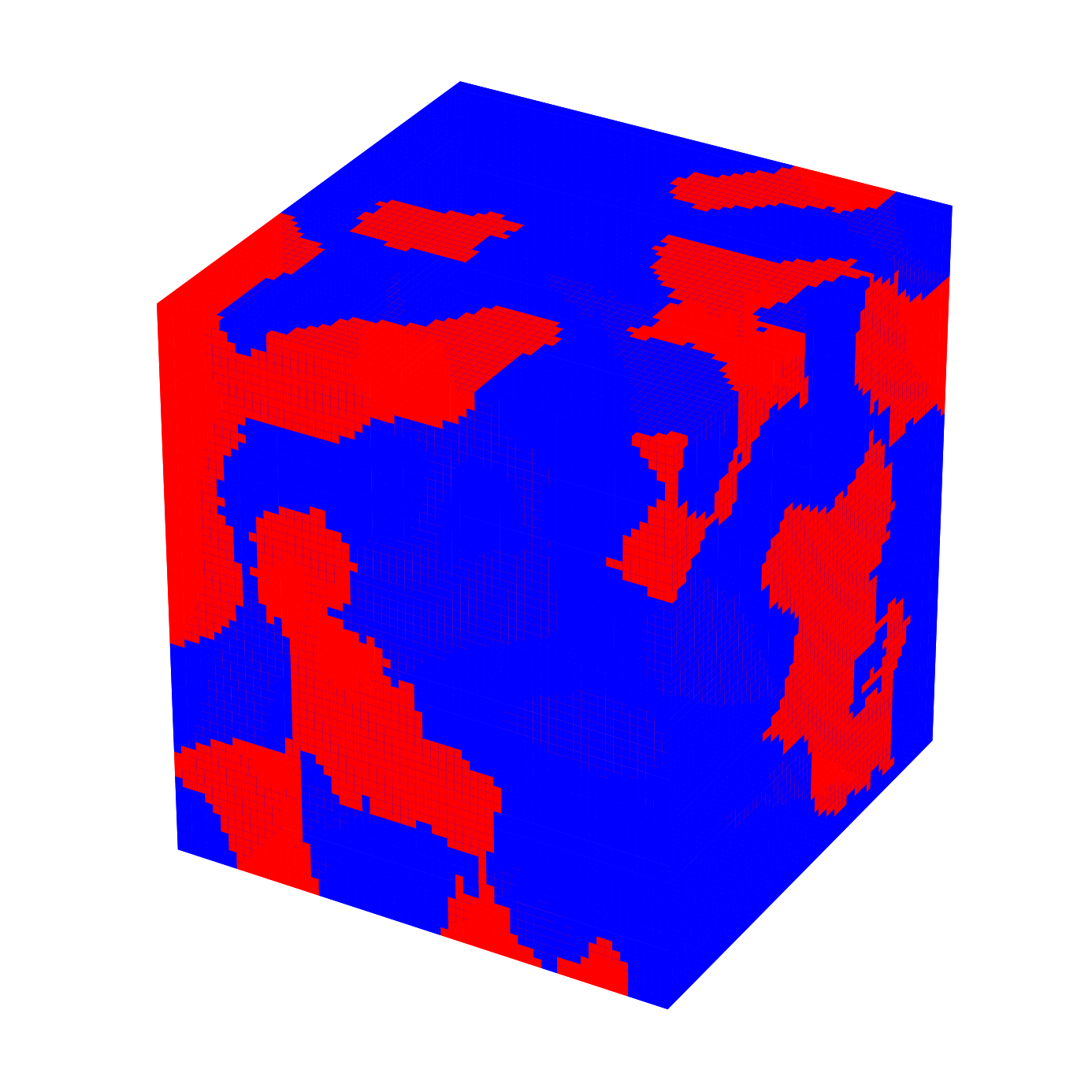}
    \end{subfigure}
    \hfill
    \begin{subfigure}{0.32\textwidth}
        \centering
         \includegraphics[width=\linewidth]{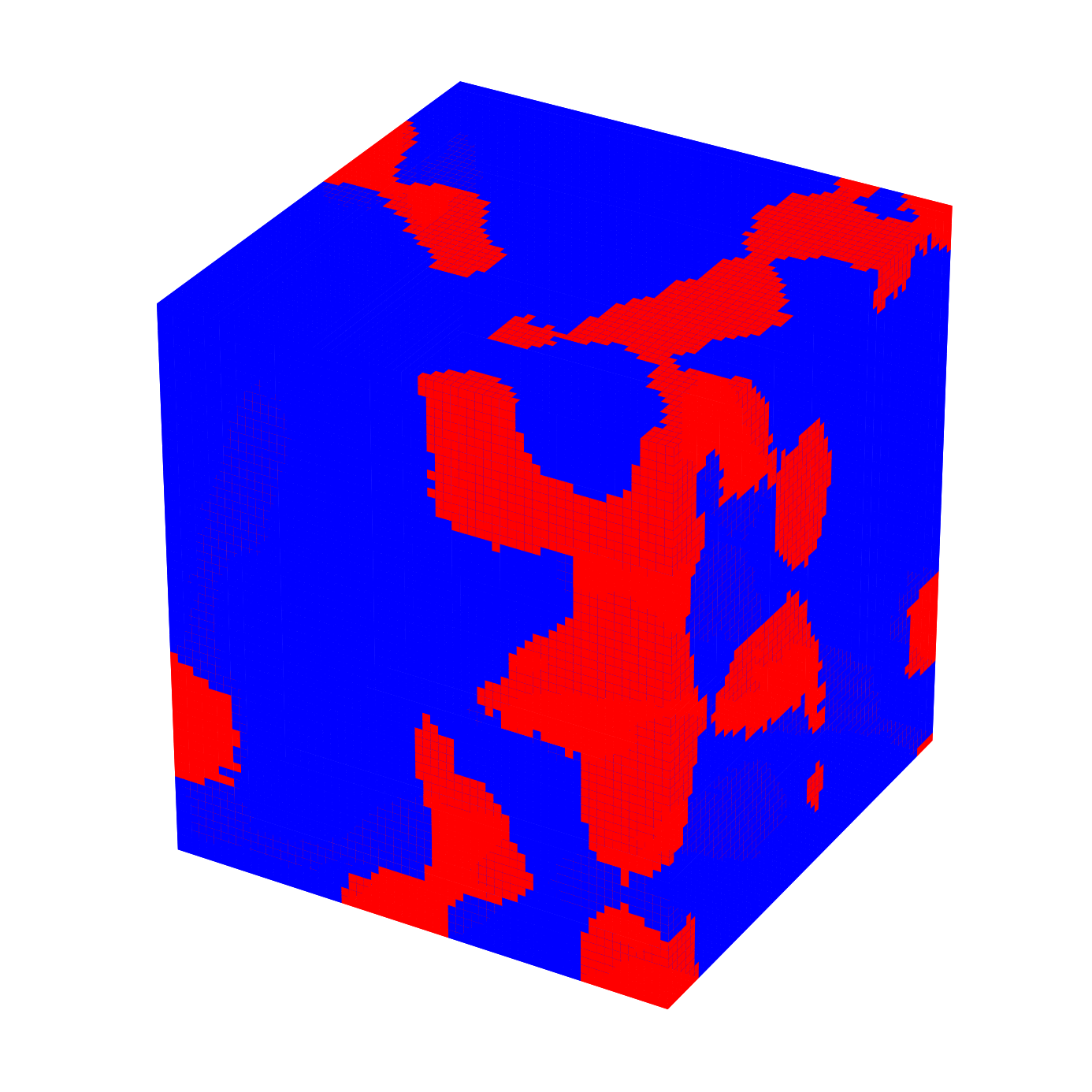}
    \end{subfigure}
    \hfill
    \begin{subfigure}{0.32\textwidth}
        \centering
         \includegraphics[width=\linewidth]{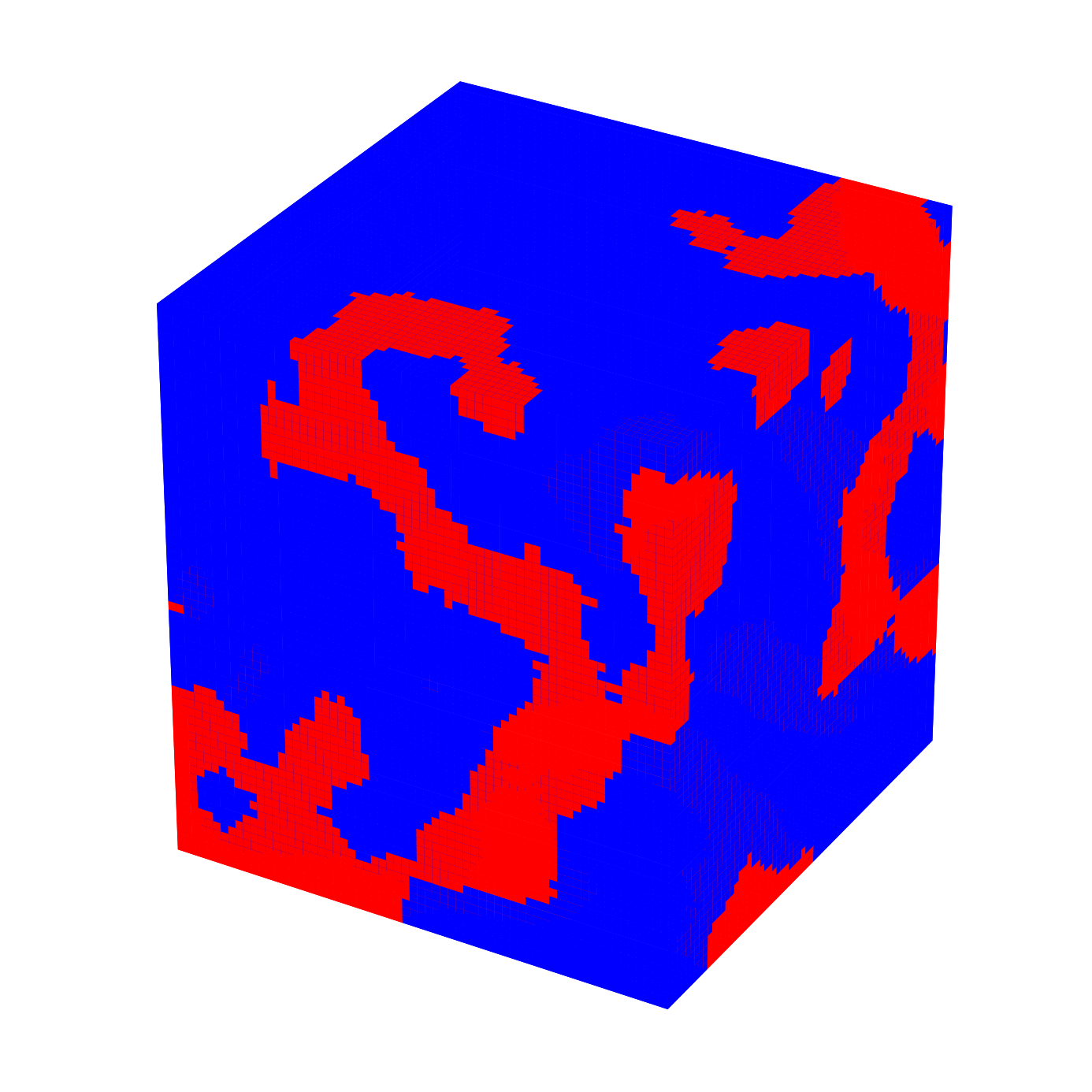}
    \end{subfigure}
    \caption{Three representative examples of the synthetically generated three-dimensional digital porous media (with $n=64$) used to train Vision Mamba are shown; phases are color-coded with blue indicating the solid grain matrix and red indicating the pore space.}
    \label{Fig1}
\end{figure*}

To test the deep-learning framework, we synthesize voxelized porous media using the truncated-Gaussian construction \cite{Lantuejoul2002Geostatistical,LeRavalec2004Conditioning}. Each sample occupies a cube of side length $L$ discretized on an $n \times n \times n$ grid, with morphology characterized by a target porosity $\phi$ (i.e., pore-volume fraction) and a spatial correlation length $\ell_c$. To construct each volumetric sample, we follow three stages. First, we generate a $64\times64\times64$ (i.e., $n=64$) scalar field of white noise by drawing samples from a standard normal distribution at every voxel. Second, we impose spatial correlation by convolving the field with a three-dimensional Gaussian kernel with a standard deviation of 5.0 and a spatial correlation length $\ell_c = 17$ voxels. Third, we rescale the smoothed field to the interval $[0,1]$ and apply a global threshold of $0.45$ so that values less than or equal to $0.45$ are labeled as pore and values greater than $0.45$ are labeled as grain, yielding a binary pore--grain medium. The selected threshold constrains the porosity to $\phi \in [0.125,\,0.200]$. In the present study, the characteristic domain length $l$ is defined as $l = n \, \Delta x$, where $\Delta x$ denotes the physical length associated with each side of a single pixel in the discretized porous medium. For all simulations, $\Delta x$ is set as $0.003 \, \text{m}$, thereby setting the spatial resolution of the computational grid. We generate 1692 samples and randomly partition them into three disjoint subsets: 1353 for training, 169 for validation, and 170 for testing. A few examples of these cubic porous media are shown in Fig. \ref{Fig1}.


Flow through each synthesized porous medium is driven by imposing a uniform streamwise pressure gradient $\Delta p/l$ in the $x$-direction. The two bounding $y$–$z$ faces are assigned no-slip conditions. Within the pore space, we compute the steady incompressible motion using a lattice Boltzmann solver \cite{LBM} that resolves the Stokes system,

\begin{equation}
\label{Eq1}
    \nabla \cdot \boldsymbol{\mathit{u}}=0,
\end{equation}

\begin{equation}
\label{Eq2}
    \nabla p - \mu \nabla^2 \boldsymbol{\mathit{u}}=\textbf{0},
\end{equation}
where $\mu$ is the dynamic viscosity and $\boldsymbol{\mathit{u}}$ and $p$ denote the velocity and pressure fields, respectively. From the converged solution, the intrinsic permeability ($k$) in the $x$-direction is obtained via Darcy's law \cite{Darcy1856Fontaines},

\begin{equation}
  \label{perm}
    k = -\frac{\mu\,U \,l}{\Delta p},
\end{equation}
with $U$ the superficial (volume-averaged) velocity evaluated over the entire sample (assigning zero velocity in solid voxels). Across the dataset, the resulting permeabilities lie within $[20\,\mathrm{mD},\,200\,\mathrm{mD}]$.

\section{Vision Mamba architecture}
\label{Sect3}

\begin{figure*}[htp]
    \centering
    \begin{subfigure}{0.85\textwidth}
        \centering
         \includegraphics[width=\linewidth]{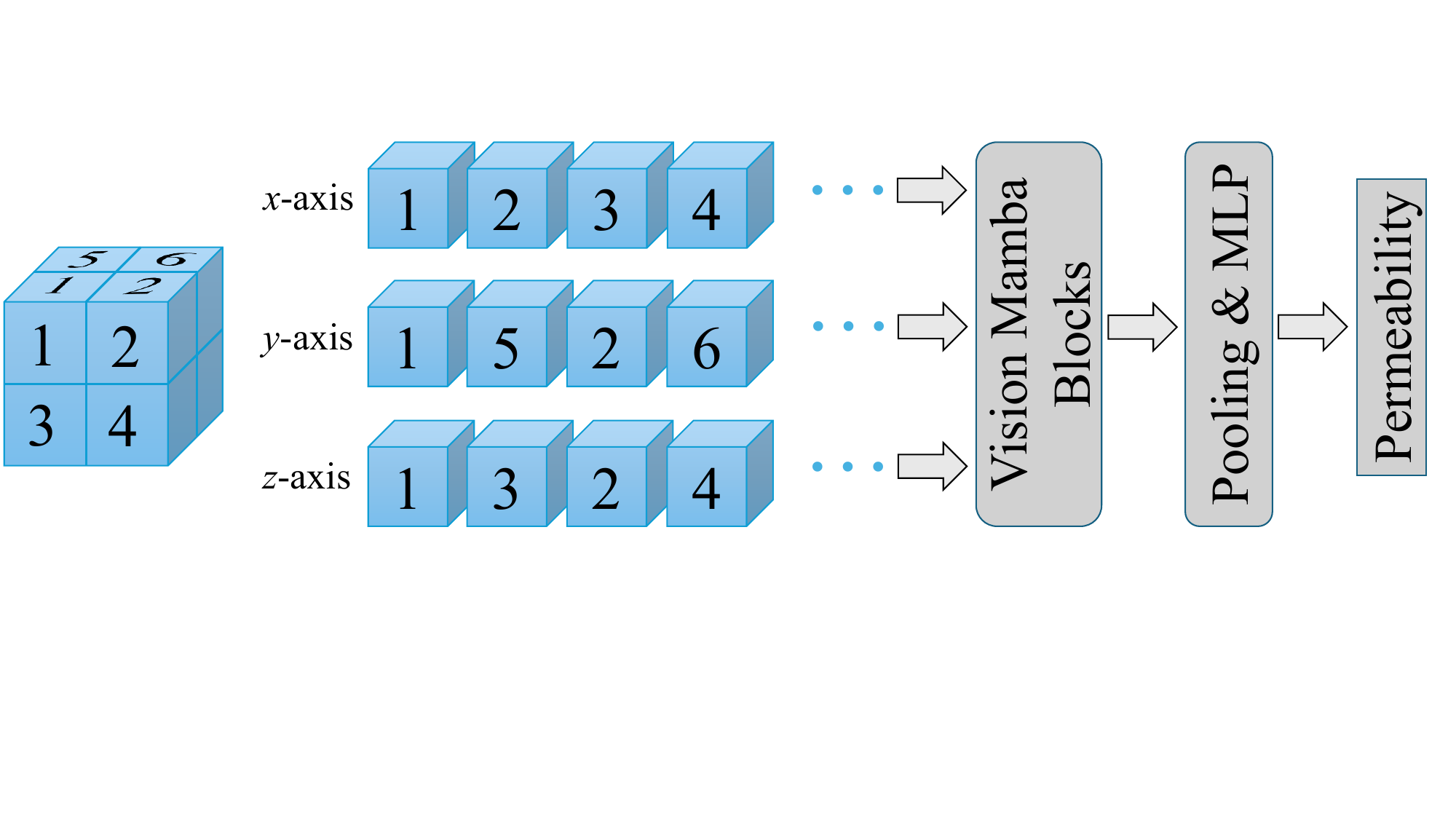}
    \end{subfigure}
    \caption{Schematic architecture of the proposed network based on Vision Mamba for deep learning of the permeability of three-dimensional porous media. If the input is a $64 \times 64 \times 64$ porous medium (i.e., $n = 64$) with a patch size of 32, there are eight subcubes, labeled 1 through 8. By scanning along the $y$-axis, the subcubes are arranged in the order 1, 5, 2, 6, 3, 7, 4, 8. Scanning along the $x$ and $z$ directions is defined similarly.}
    \label{FigShematic}
\end{figure*}

In this section, we describe the architecture of the proposed neural network, whose core is Vision Mamba, a selective state-space model, adapted to predict permeability from voxelized porous media. Figure \ref{FigShematic} illustrates the schematic of Vision Mamba, which serves as the core of the proposed neural network, with the 3D porous medium cube as the input. We split the cube into non-overlapping 3D patches, and each patch is embedded into a token. A stack of Vision Mamba blocks scans the embedded tokens along the depth, height, and width axes. Each scan performs bidirectional state updates through forward and backward recurrences. Finally, we take a global average over space and use a linear layer to output a single permeability value.

\subsection{Input and patchification}
The input to the network is a batch of generated porous media, i.e., a batch of cubes, which mathematically can be shown by $X\in\mathbb{R}^{\mathcal{B}\times 1\times D\times H\times W}$, where $\mathcal{B}$ is the batch size and $D$, $H$, and $W$ are spatial dimensions. Next, we apply a patchification operator. Patchification uses a single 3D convolution with kernel size and stride equal to the patch size. In this setup, the patchification produces an $D' \times H' \times W'$ grid of patch tokens, each with $\mathcal{C}$ channels. Consequently, the output of the patchification operator is the token grid $z_\text{tok}\in\mathbb{R}^{\mathcal{B}\times \mathcal{C} \times D' \times H' \times W'}$.

\subsection{Vision Mamba block} 
The token grid $z_\text{tok}$ (obtained from the previous step) serves as the input to Vision Mamba. To elaborate on the process within Vision Mamba, we describe it in three stages: token-wise parameter generation, selective scanning along each axis, and axis fusion with residual connections.

\subsubsection{Token-wise parameter generation}
 In the next step, a $1 \times 1 \times 1$ convolution reads $z_\text{tok}\in\mathbb{R}^{\mathcal{B}\times \mathcal{C} \times D' \times H' \times W'}$ and produces five fields of size $\mathcal{B}\times \mathcal{C} \times D' \times H' \times W'$: input gate ($g_\text{in}$), output gate ($g_\text{out}$), and two state-space coefficients $B$ and $C$, as well as positive step size $\Delta$. Moreover, we create a learnable vector $A \in \mathbb{R}^{\mathcal{C}}$ and a skip vector $D_{\text{skip}} \in \mathbb{R}^{\mathcal{C}}$. In addition, we define $u$ as
\begin{equation}
u = g_\text{in} \odot z_\text{tok},
\end{equation}
where $\odot$ denotes elementwise product and thus $u\in\mathbb{R}^{\mathcal{B}\times \mathcal{C} \times D' \times H' \times W'}$. $A_\text{+}$ is introduced and computed by applying the elementwise softplus function to the vector $A$. The softplus function is defined as
\begin{equation}
\sigma(\lambda)=\ln\!\big(1+e^{\lambda}\big).
\end{equation}
Note that although $A_\text{+} \in \mathbb{R}^{\mathcal{C}}$, it is treated as $A_\text{+} \in \mathbb{R}^{1 \times \mathcal{C} \times 1 \times 1 \times 1}$ in practice from a software engineering perspective. Next, $\alpha$ is introduced and computed as
\begin{equation}
    \alpha=\exp\!\big(-\,A_\text{+}\odot \Delta\big), 
\end{equation}
where $\alpha\in\mathbb{R}^{\mathcal{B}\times \mathcal{C} \times D' \times H' \times W'}$.

\subsubsection{Selective scan per axis} 
For each spatial axis $\eta\in\{D',H',W'\}$, the token grid $z_\text{tok}$ is viewed as a collection of length-$L_\eta$ sequences by treating that axis as an ordered time dimension and flattening the remaining indices into independent sequences. The forward selective state-space scan along axis $\eta$ updates a channelwise hidden state $s$ and produces an output $y^{\text{fwd}}$ via

\begin{equation}\label{eq:ssm_state}
s_t=\alpha_t\odot s_{t-1}+B_t\odot u_t,
\end{equation}
\begin{equation}\label{eq:ssm_out}
y_t^{\text{fwd}}=C_t\odot s_t+D_\text{skip}\odot u_t,
\end{equation}
from $t = 0$ to $t=L_\eta-1$. Note that the subscript $t$ added to the components $s$, $\alpha$, $B$, $u$, $C$, and $y^{\text{fwd}}$ (i.e., $s_t$, $\alpha_t$, $B_t$, $u_t$, $C_t$, and $y^{\text{fwd}}_t$) indicates that these components are reshaped to $N_\text{seq} \times L \times \mathcal{C}$, and therefore $\{s_t, \alpha_t, B_t, u_t, C_t, y^{\text{fwd}}_t\} \in \mathbb{R}^{N_\text{seq} \times \mathcal{C}}$. The value of $N_\text{seq}$ depends on the scanning axis. For example, when scanning along the $D'$ axis, $N_\text{seq} = \mathcal{B} \times H' \times W'$. Similarly, $N_\text{seq}$ is computed when the other two axes are scanned. Finally, a corresponding backward scan runs from $t = L_\eta$ to $t = 1$, producing $y^{\text{bwd}}$.

\subsubsection{Axis fusion and residuals}

The two directions are fused to remove directional bias,  
\begin{equation}\label{eq:bidir}
    \hat{y} = \tfrac{1}{2}\big(y^{\text{fwd}} + y^{\text{bwd}}\big),
\end{equation}
and the result is then gated at the output:  
\begin{equation}\label{eq:axis_gate}
    y_\eta = g_{\text{out}} \odot \hat{y}.
\end{equation}
The outputs along the three axes are subsequently averaged to obtain the final output:  
\begin{equation}\label{eq:axis_fuse}
    y_\text{tok} = \frac{y_{D'} + y_{H'} + y_{W'}}{3}.
\end{equation}
The Vision Mamba block employs a residual connection and a pointwise MLP to mix channels after the selective scan, formulated as  
\begin{equation}
    z^{+} = z_\text{tok} + y_\text{tok},
\end{equation}
\begin{equation}
    z_\text{out} = z^{+} + \mathcal{M}(z^{+}),
\end{equation}
where $\mathcal{M}$ denotes a pointwise MLP implemented using $1 \times 1 \times 1$ convolutions with the Gaussian error linear unit ($\mathcal{G}$) activation function, defined as
\begin{equation}
\mathcal{G}(\lambda) = \frac{\lambda}{2} ( 1 + \text{erf} (2^{-0.5} \lambda)).
\end{equation}



\subsection{Global pooling and head}

After the Vision Mamba block, a global average pooling across spatial dimensions and a linear projection to a scalar is applied. If $h$ denotes the globally pooled representation, the predicted permeability $\hat{k}$ is computed as
\begin{equation}\label{eq:head_h}
h=\operatorname{mean}(z_\text{out}),
\end{equation}
\begin{equation}\label{eq:head_y}
\hat{k}=w^{\top}h+b,
\end{equation}
where $w$ is the weight vector and $b$ is a scalar bias.

Note that we described the architecture of a single Vision Mamba block. Multiple blocks can be stacked sequentially to construct deeper networks. For further details on the underlying methodology, we refer readers to the original Mamba formulation \cite{gu2023mamba} and its vision-oriented adaptation in Vision Mamba \cite{zhu2024vision}. Moreover, implementation-specific details are documented in our openly available GitHub repository (see the Data Availability part at the end of the article), which includes extensive inline comments.

\section{Parameter setup and training}
\label{Sect4}

Since $n=64$, the parameters $D$, $W$, and $H$ are set to $64$. We set a patch size of $8$ voxels in the patchification operator on $64^3$ inputs, producing an $8\times 8\times 8$ grid of $512$ tokens with embedding width $\mathcal{C}=64$. Since the input dimensions are $64$ and the patch size is $8$, it follows that $D'$, $W'$, and $H'$ are $8$ (because $64/8=8$). Moreover, because $n=64$ and the patch size equals $8$, it is concluded that the sequence length along each axis is $L_\eta=8$ (the number of pixels divided by the patch size). Additionally, we set the block depth $N_{\text{block}}$ to 3. The block depth of 3 ($N_{\text{block}}=3$) means that three Vision Mamba blocks are stacked sequentially after the patch-embedding stem, each applying axiswise bidirectional selective scans and residual pointwise mixing to the output of the preceding block before the global pooling and linear regression head. In Sect. \ref{Sect53}, we report a series of ablation studies that systematically examine how these hyperparameters affect the predictive performance of the model.

Training uses mean-squared error on a min–max normalized target. Let $k_{\min}$ and $k_{\max}$ be the minimum and maximum permeability values computed on the training split. The normalized target is
\begin{equation}\label{eq:norm}
\widetilde{k}=\frac{k-k_{\min}}{k_{\max}-k_{\min}}.
\end{equation}
The loss over a batch is
\begin{equation}\label{eq:loss}
\mathcal{L}=\frac{1}{\mathcal{B}}\sum_{i=1}^{\mathcal{B}}\big(\hat{k}_i-\widetilde{k}_i\big)^2.
\end{equation}
At evaluation time, predictions are mapped back to physical units by the inverse of Eq. \ref{eq:norm}. This normalization stabilizes optimization without imposing a hard output range; the regression head remains unconstrained and learns to match the normalized scale.

Model training proceeds via stochastic, mini-batch gradient optimization by adopting the Adam optimizer \cite{kingma2014adam} with a constant learning rate of 0.001, and using mini-batches of 128 samples (i.e., $\mathcal{B}=128$) for each parameter update \cite{Goodfellow2016}. To avoid overfitting, model performance is continuously monitored on a held-out validation set throughout training. Convergence is typically achieved within approximately 300 epochs, at which point the final optimized model is selected. All experiments are executed on a single NVIDIA A100 (SXM4) GPU equipped with 80 GB of memory.

\begin{figure*}[htp]
    \centering
    \begin{subfigure}{0.47\textwidth}
        \centering
         \includegraphics[width=\linewidth]{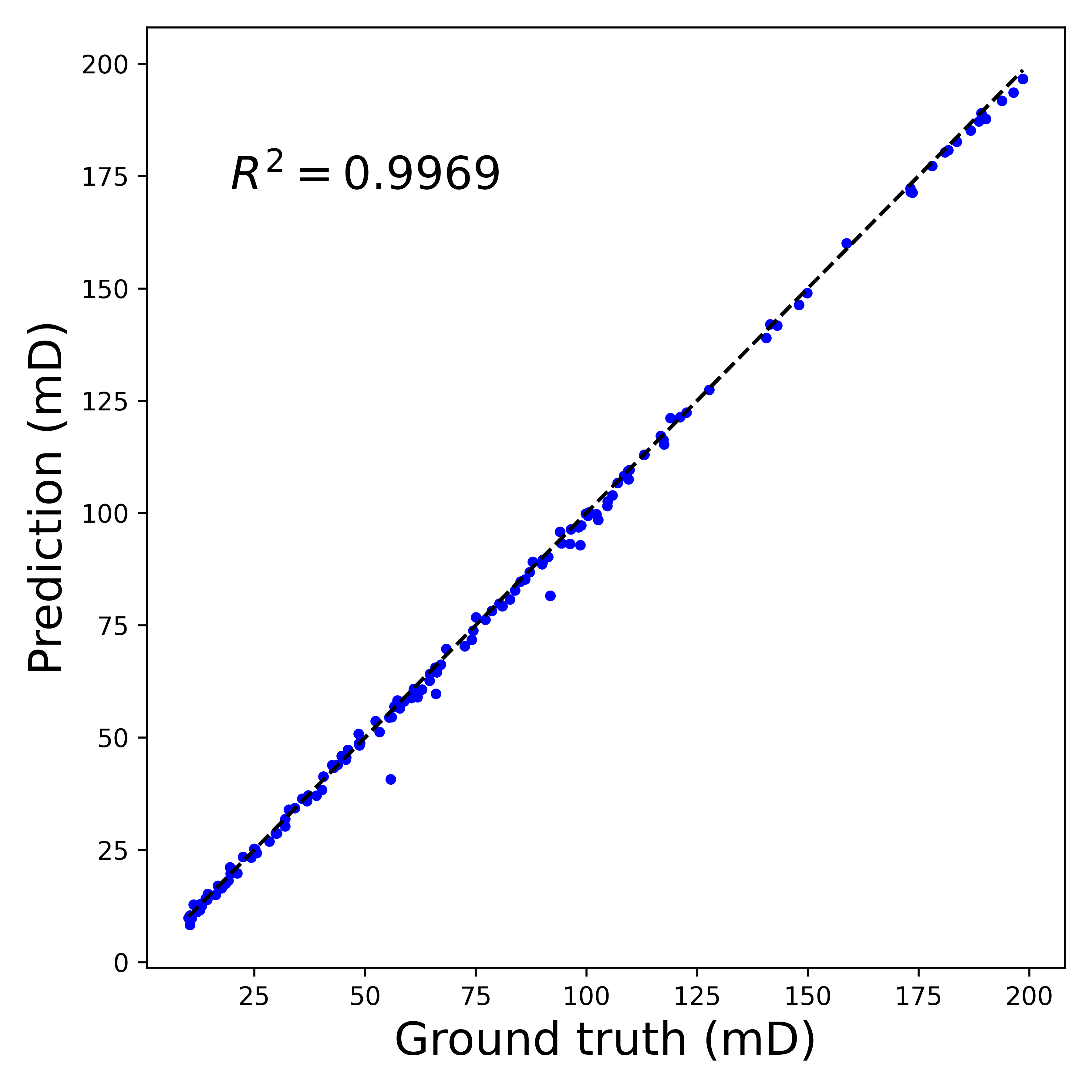}
    \end{subfigure}
    \hfill
    \begin{subfigure}{0.47\textwidth}
        \centering
         \includegraphics[width=\linewidth]{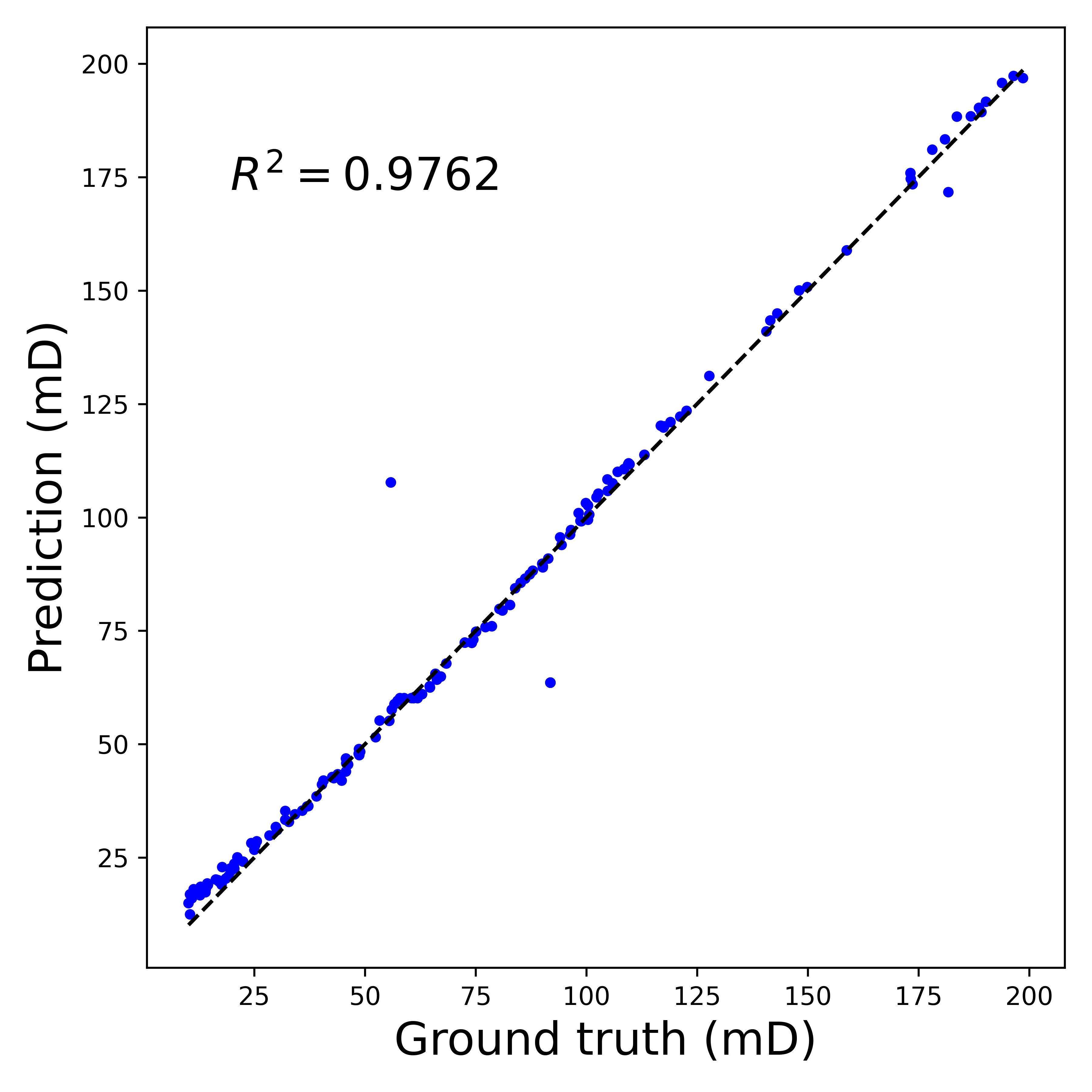}
    \end{subfigure}
    \caption{Comparison of Vision Mamba and CNN performance using the $R^2$ score (left: Vision Mamba; right: CNN)}
    \label{Fig2}
\end{figure*}

\section{Results and discussion}
\label{Sect5}

\subsection{General analysis}
\label{Sect51}

To assess predictive accuracy for permeability, we employ the coefficient of determination, $R^2$. For a test set comprising $P$ samples with ground-truth permeabilities $k_i$ and corresponding predictions $\tilde{k}_i$, and denoting by $\bar{k}=\frac{1}{P}\sum_{i=1}^{P}k_i$ the empirical mean of the ground truth, $R^2$ is defined as
\begin{equation}
R^2 \;=\; 1 \;-\; \frac{\sum_{i=1}^{P}\big(k_i-\tilde{k}_i\big)^2}{\sum_{i=1}^{P}\big(k_i-\bar{k}\big)^2}.
\end{equation}
Note that negative values of $R^2$ imply performance inferior to the trivial predictor $k_i=\bar{k}$. We also report the root-mean-square error (RMSE), defined as
\begin{equation}
    \mathrm{RMSE} = \sqrt{ \frac{1}{P} \sum_{i=1}^{P} \big(k_i - \tilde{k}_i\big)^2 }.
\end{equation}
We further report the maximum relative error over the test set by computing $\max \Big \{\frac{\lvert k_i - \tilde{k}_i \rvert}{\lvert k_i \rvert} \Big \}_{i=1}^P$. The minimum relative error is similarly defined.

\begin{table}[h]
 \centering
 \small
\caption{$R^2$ score, root mean square error, and minimum/maximum relative errors of the test set (170 samples) for the comparison between Vision Mamba and CNN models. The batch size ($\mathcal{B}$) for both models is set to 128. We set $N_\text{block}=3$ and as well as the patch size of 8 in the Vision Mamba model.}\label{Table0}
\begin{tabular}{lllllll}
\toprule
  & Vision Mamba & CNN   \\
\midrule
 $R^2$ score &  0.9969 &  0.9762 \\
 Root mean square error (mD) & 2.6939  & 7.5054 \\
 Minimum relative error & 0.0003 &  0.0004 \\
 Maximum relative error & 0.2708 & 0.9312 \\
 Training time per epoch (s) & 3.0 & 1.6 \\
 Number of trainable parameters & 195841 &  2582369 \\
\bottomrule
\end{tabular}
\end{table}

The performance and error analysis of the Vision Mamba model in predicting the permeability of the test set (170 porous media) are summarized in Table \ref{Table0}. As reported, the $R^2$ score is 0.9969 and the root mean square error is 2.6939 mD. The maximum and minimum relative errors are 0.2708 and 0.0003, respectively. The left panel of Fig. \ref{Fig2} further illustrates the predicted versus ground-truth permeability for all samples in the test set, highlighting the results for individual porous media. These findings demonstrate the successful training and accurate predictive capability of the proposed Vision Mamba–based neural network for applications to three-dimensional porous media.

\subsection{Comparison between Vision Mamba and CNNs}
\label{Sect5compCNN}

The next step is to compare the performance of Vision Mamba with that of a CNN. For completeness, we briefly outline the CNN architecture used in this study. Specifically, we employ the same CNN model that was adopted in our previous work published in 2021 \cite{kashefi2021PointNetPorousMedia}. In simple terms, the CNN model consists of an encoder and a decoder. In the encoder, convolutional channels start at 16 and double at each stage. Downsampling is done with stride-2 convolutions without padding. There are no pooling layers in the encoder. We use \(2\times 2\times 2\) kernels, except for the last layer of the encoder, which uses a \(1\times 1\times 1\) kernel. This final layer produces a single global latent vector of length \(1024\). This latent vector is then passed to a decoder, implemented as a multilayer perceptron (MLP) with three layers of sizes 512, 256, and 1, respectively, which is used to predict the permeability. In both the encoder and decoder, Rectified Linear Unit (ReLU) activation function (see Eq. 7 in Ref. \cite{kashefi2021PointNetPorousMedia} for the mathematical expression of this function) is applied as the activation function after each layer except the final layer, which has no activation. Batch normalization \cite{ioffe2015batch} is applied after each layer. In the decoder, dropout \cite{srivastava2014dropout} with a probability of 0.7 is used. Similar to the Vision Mamba model, the loss function is the mean squared error (Eq. \ref{eq:loss}). For additional background and implementation details on CNNs for permeability prediction in porous media, see Ref. \cite{kashefi2021PointNetPorousMedia}.

The performance and error analysis of the predicted permeability values of the test set using the CNN model are listed in Table \ref{Table0}, with the corresponding results illustrated in the right panel of Fig. \ref{Fig2} for the $R^2$ score. In comparison with Vision Mamba, the CNN yields a lower $R^2$ score (0.9762 vs. 0.9969), a higher root mean square error (7.5054 mD vs. 2.6939 mD), a higher minimum relative error (0.0004 vs. 0.0003), and a higher maximum relative error (0.9312 vs. 0.2708). It is important to emphasize that our focus here is not solely on showing that Vision Mamba consistently outperforms CNN in terms of prediction accuracy of porous media permeability. In fact, we ensured that the CNN model was optimized to achieve its best possible performance. Nevertheless, as discussed earlier, CNN still achieves lower $R^2$ scores compared to Vision Mamba. Instead, our comparison primarily concerns the training time and the number of trainable parameters, as summarized in Table \ref{Table0}. The training time (per epoch) of Vision Mamba is approximately 1.875 times longer than that of CNN. This can be attributed to the sequential nature of Vision Mamba, which converts each three-dimensional porous medium into a sequence of patches and processes them serially. In contrast, CNNs process three-dimensional porous media through multiple channels in parallel, where the number of channels typically increases and the kernel size decreases at deeper layers, leading to faster training. However, this parallelization comes at the cost of requiring more trainable parameters and higher GPU memory consumption. Based on Table \ref{Table0}, the number of trainable parameters in the CNN model is 2582369, whereas Vision Mamba requires only 195841 parameters, approximately a 13.2-fold reduction, which is a substantial difference.

\subsection{Comparison between Vision Mamba and ViT}
\label{Sect5compViT}

\begin{figure*}[htp]
    \centering
    \begin{subfigure}{0.70\textwidth}
        \centering
         \includegraphics[width=\linewidth]{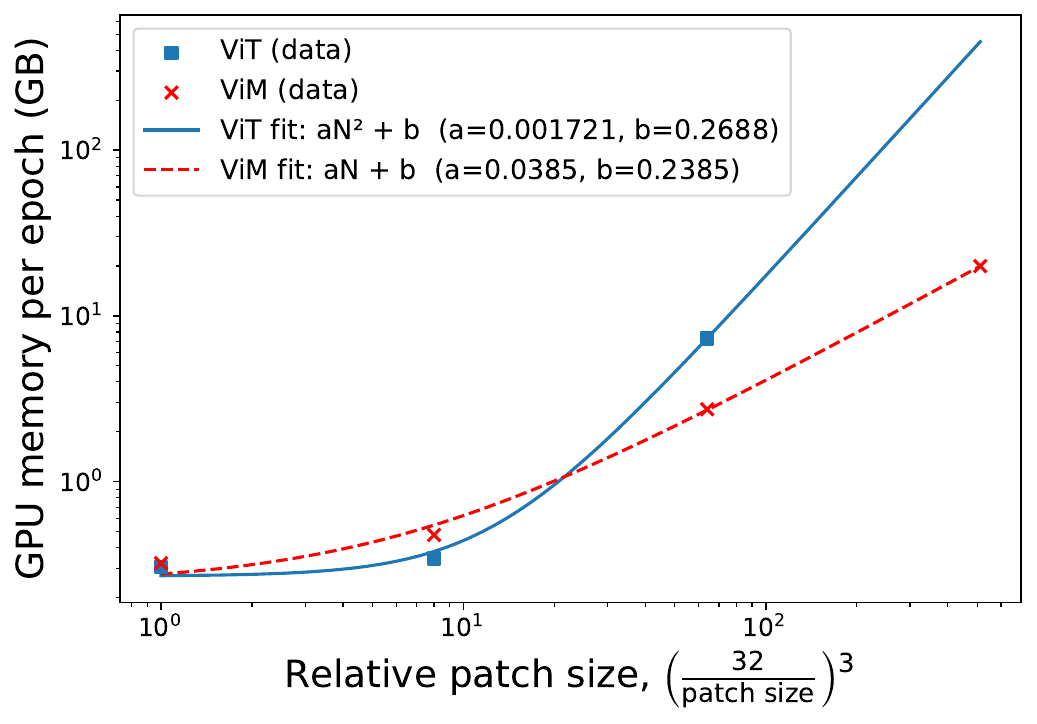}
    \end{subfigure}
  
    \caption{GPU memory (per epoch) scaling versus relative patch size (i.e., relative token size) for Vision Mamba (ViM) and ViT on log--log axes. Least-squares fits reveal linear scaling for Vision Mamba and quadratic scaling for ViT.}
    \label{FigScale}
\end{figure*}

This subsection compares the proposed model, based on Vision Mamba, with ViT for permeability prediction of porous media. A brief summary of the ViT architecture is provided at the end of this subsection; here, we focus on the results. In each machine learning experiment, the number of trainable parameters between the two models is matched as closely as possible. Once the initial ViT design is established, the only variable across experiments is the patch size (i.e., token size). The Vision Mamba configuration follows that described previously, except that the patch size is varied. For both models, the batch size is fixed at 128 ($\mathcal{B}=128$). Other hyperparameters and training procedures are selected to achieve the best performance, and early stopping is applied to mitigate overfitting. Table \ref{TableMambaViT} reports the results for patch sizes 4, 8, 16, and 32. In both models, reducing the patch size decreases the number of trainable parameters but increases GPU memory requirements. The character of this increase distinguishes Vision Mamba from ViT. Although the number of trainable parameters remains nearly constant in both models as patch size decreases, GPU memory usage grows at different rates. For ViT, the increase is so pronounced that a model with patch size 4 cannot be executed on an 80-GB NVIDIA A100 (SXM4) GPU, resulting in job failure, as reported in Table \ref{TableMambaViT}. Figure \ref{FigScale} shows the GPU memory usage per epoch as a function of patch size. In this plot, the patch size is normalized by the largest patch size (i.e., 32 in the current case). As shown in Fig. \ref{FigScale}, the required GPU memory per epoch increases linearly with decreasing patch size in Vision Mamba, whereas it increases quadratically with decreasing patch size in ViT. Our experimental results on three-dimensional porous media (i.e., a specific 3D image) confirm the theoretical design of Vision Mamba and ViT in terms of linear and quadratic scaling with token size. As illustrated in Fig. \ref{FigScale}, we apply a least-squares fit to derive linear and quadratic equations describing the experimental GPU memory requirements. Based on these equations, it is predicted that at a patch size of 4, the required GPU memory per epoch for ViT would be approximately 451 GB, which explains why this machine learning experiment could not be executed on our 80-GB GPU.

\begin{figure*}[htp]
    \centering
    \centerline{ViM, $\text{patch size}=8$ \hspace{1.5cm} ViM, $\text{patch size}=16$ \hspace{1.4cm} ViM, $\text{patch size}=32$}
    \begin{subfigure}{0.32\textwidth}
        \centering
         \includegraphics[width=\linewidth]{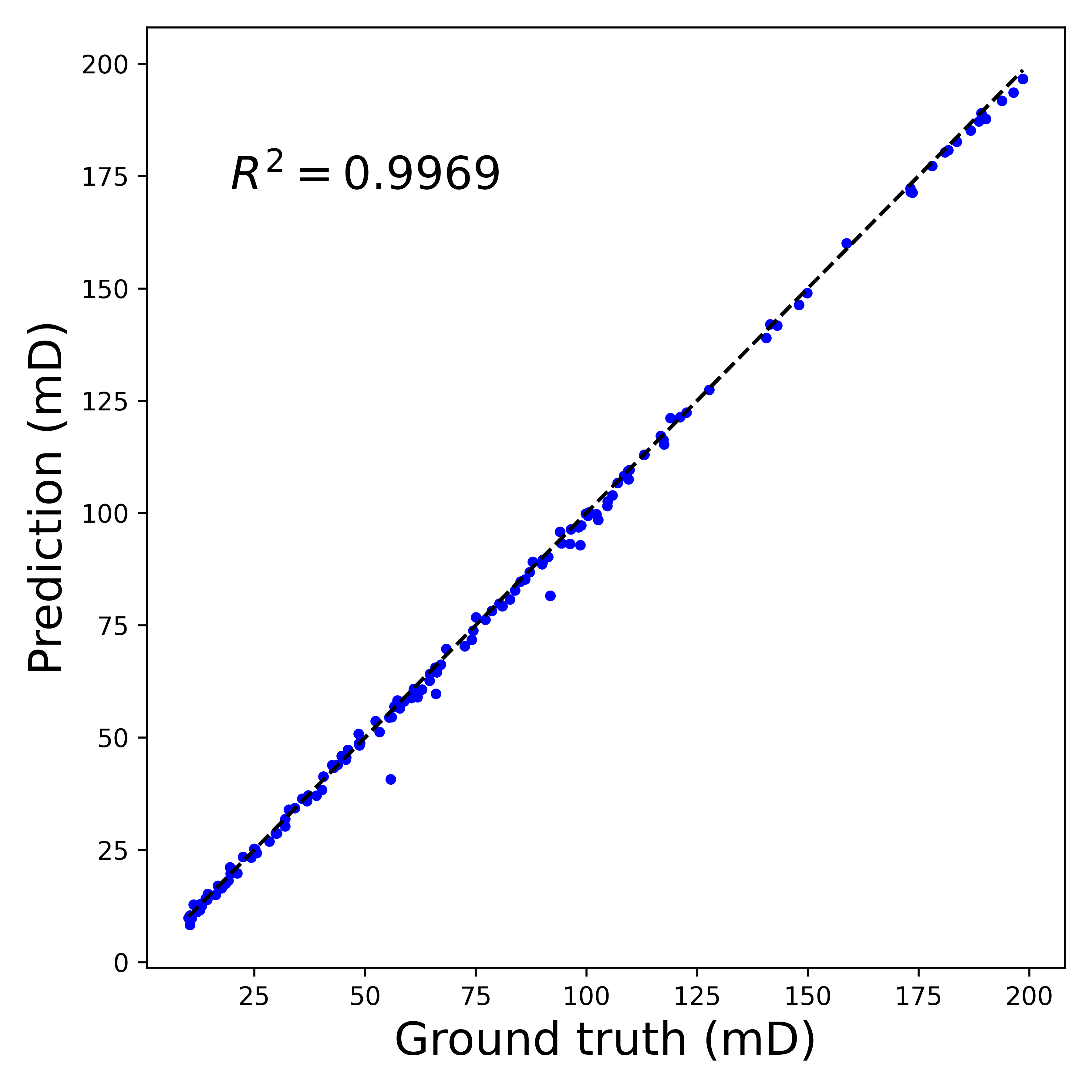}
    \end{subfigure}
    \hfill
    \begin{subfigure}{0.32\textwidth}
        \centering
         \includegraphics[width=\linewidth]{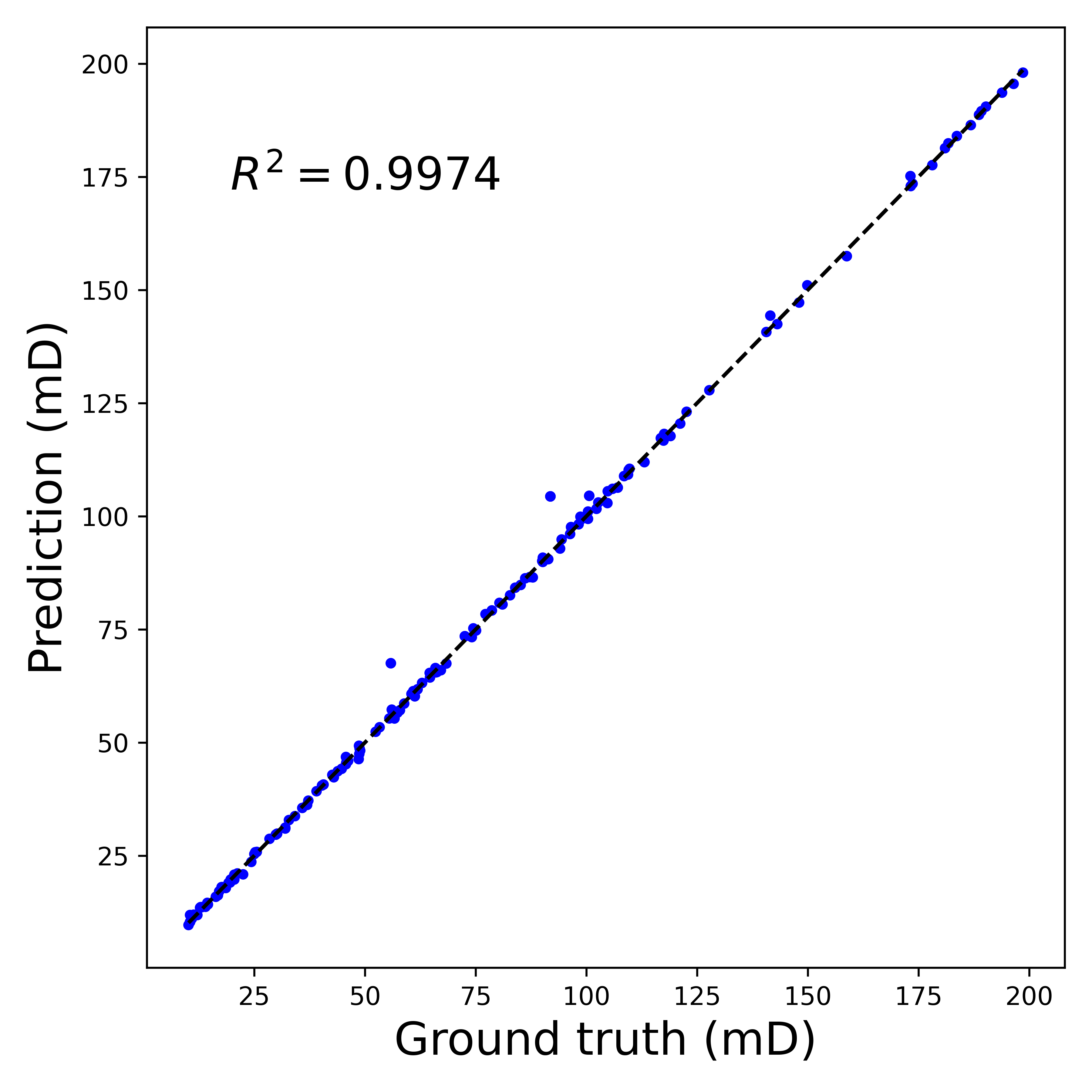}
    \end{subfigure}
     \hfill
    \begin{subfigure}{0.32\textwidth}
        \centering
         \includegraphics[width=\linewidth]{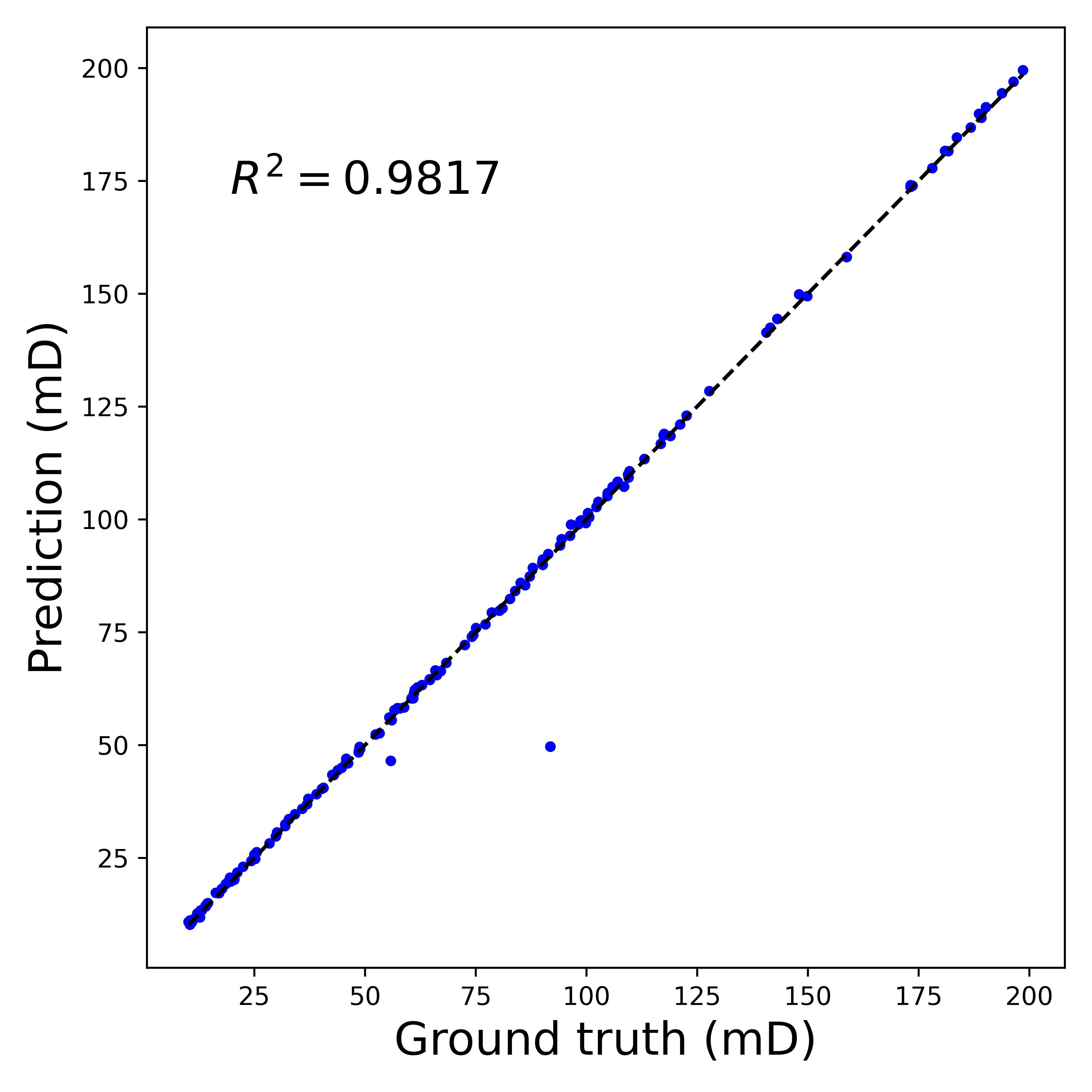}
    \end{subfigure}

    \centerline{ViT, $\text{patch size}=8$ \hspace{1.6cm} ViT, $\text{patch size}=16$ \hspace{1.4cm} ViT, $\text{patch size}=32$}
    \begin{subfigure}{0.32\textwidth}
        \centering
         \includegraphics[width=\linewidth]{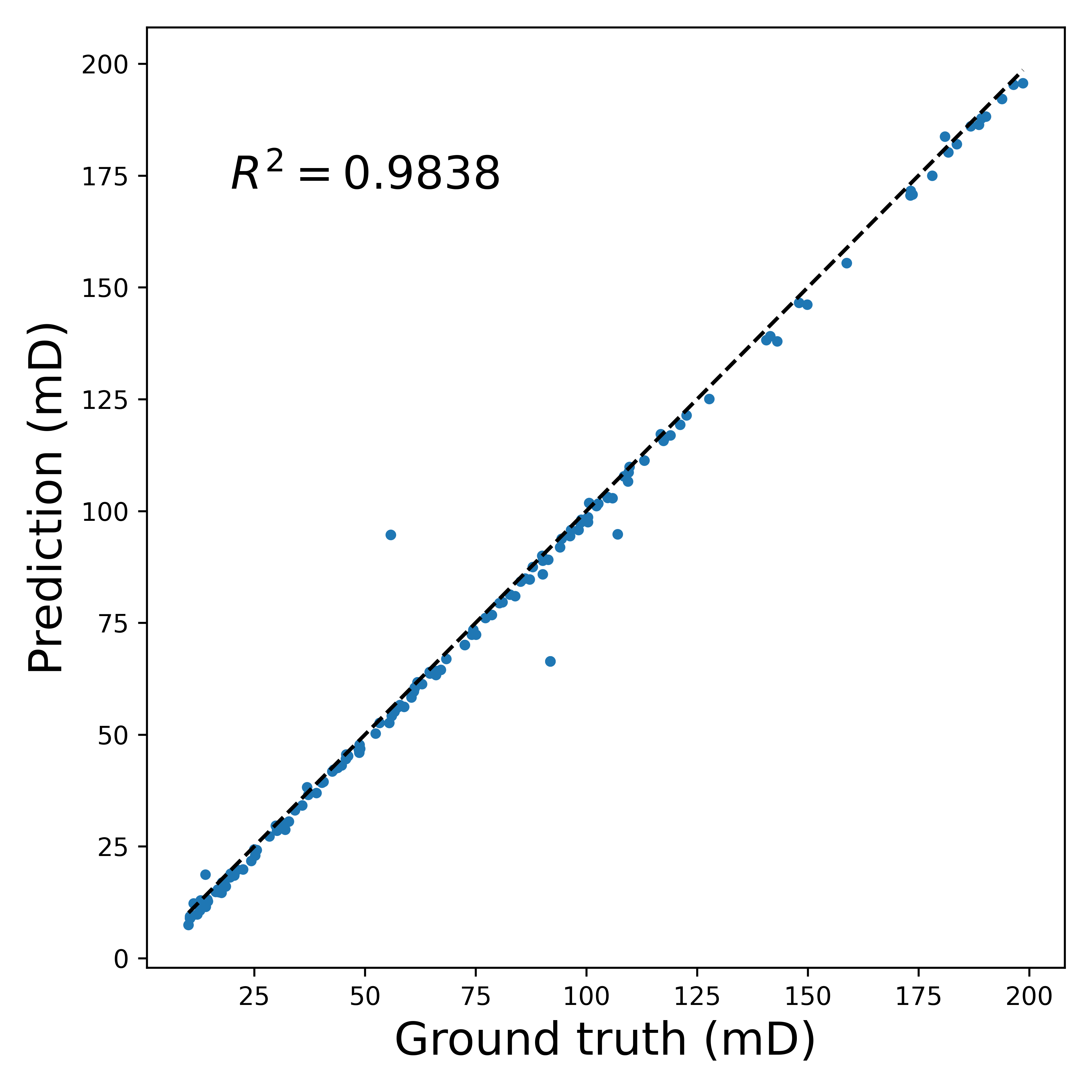}
    \end{subfigure}
    \hfill
    \begin{subfigure}{0.32\textwidth}
        \centering
         \includegraphics[width=\linewidth]{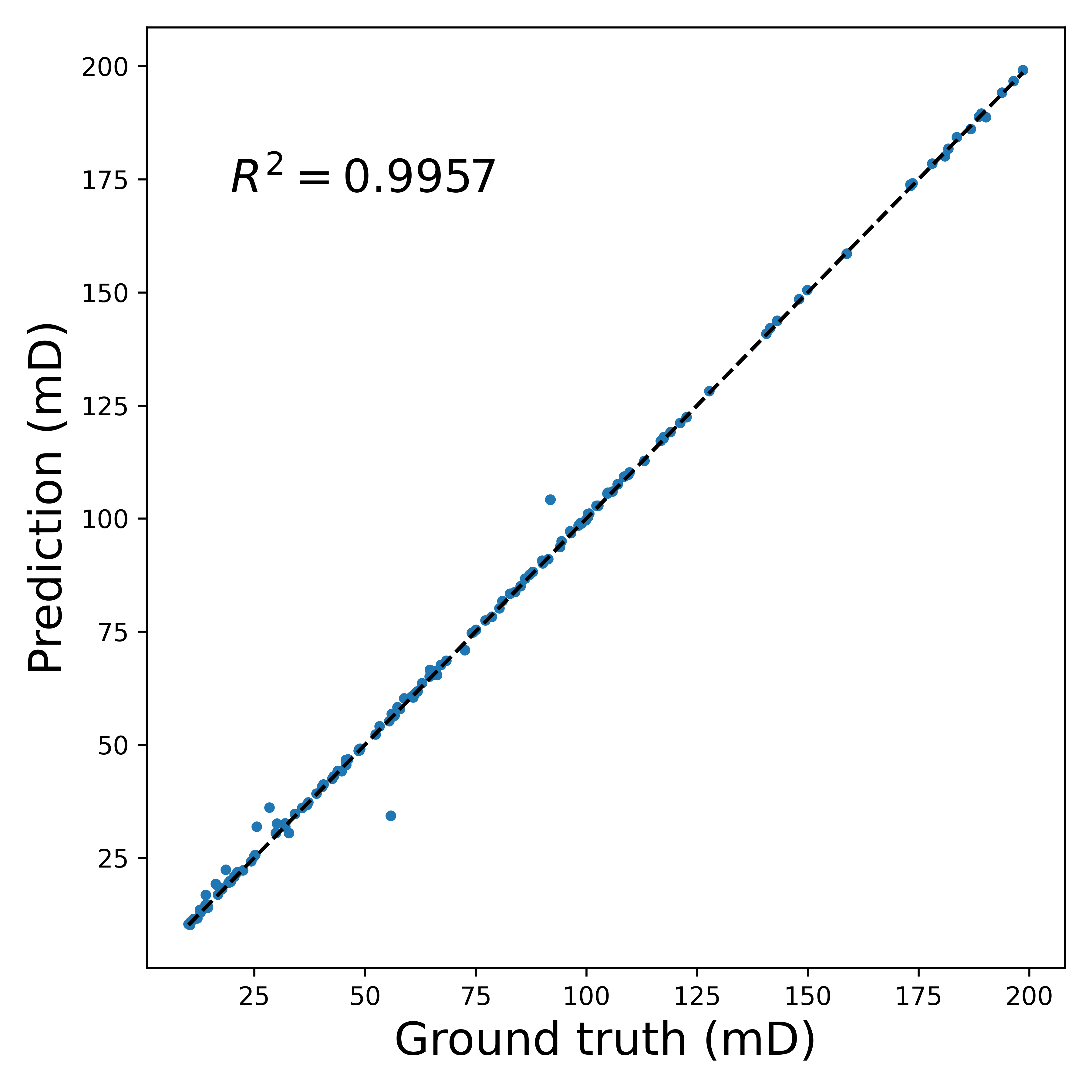}
    \end{subfigure}
     \hfill
    \begin{subfigure}{0.32\textwidth}
        \centering
         \includegraphics[width=\linewidth]{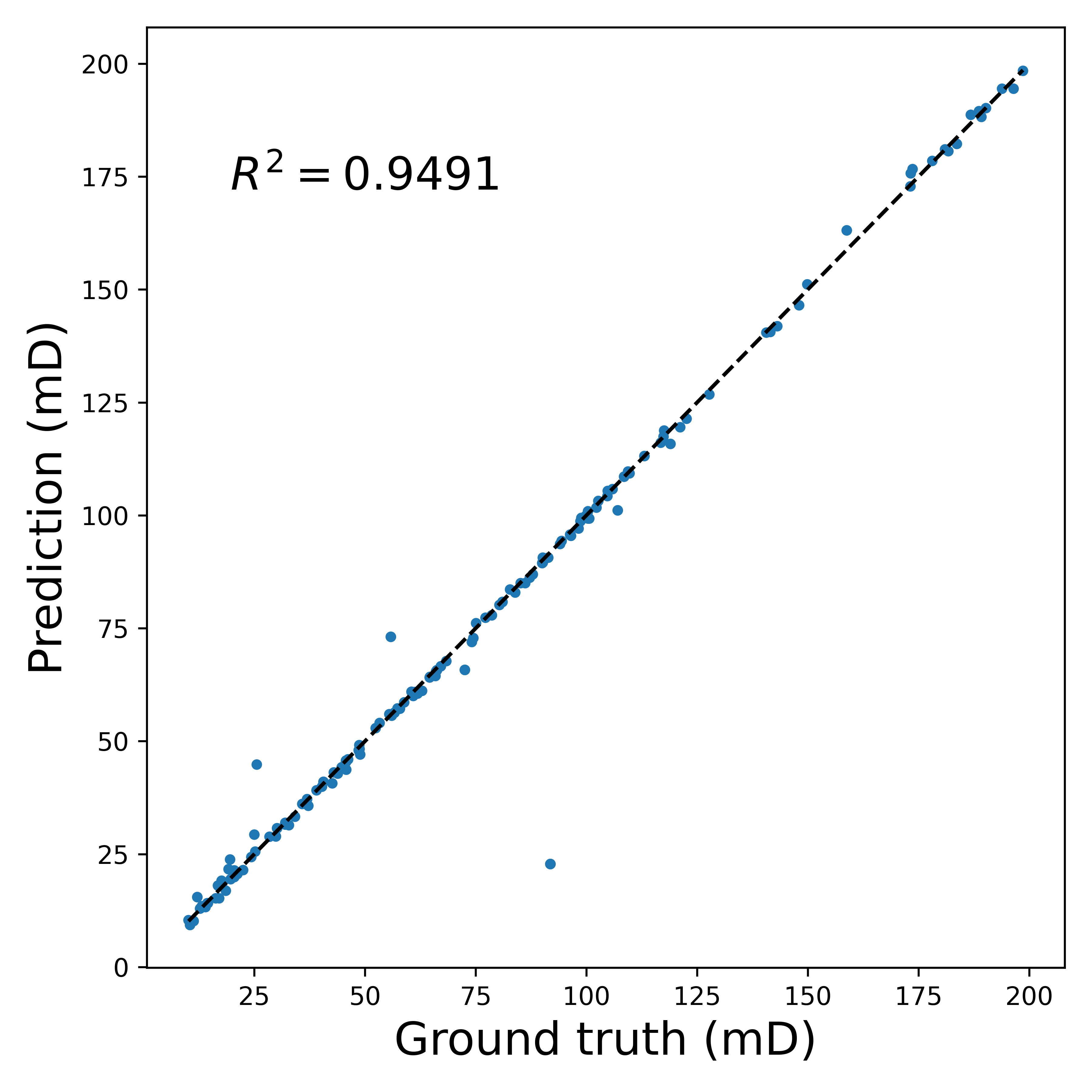}
    \end{subfigure}
    \caption{Comparison of Vision Mamba (ViM) and ViT performance based on the $R^2$ score. The first row corresponds to Vision Mamba (ViM), and the second row corresponds to ViT.}
    \label{Fig200}
\end{figure*}

Figure \ref{Fig200} presents the permeability predictions versus the ground truth for patch sizes 8, 16, and 32 using the Vision Mamba and ViT models. Based on the results reported in Table \ref{TableMambaViT} and the visualizations in Fig. \ref{Fig200}, it can be concluded that very large patch sizes reduce the accuracy of permeability prediction. This effect is more pronounced for ViT, which attains an $R^2$ score of 0.9491 at a patch size of 32, whereas Vision Mamba maintains an $R^2$ score of 0.9817 at the same patch size. Overall, Vision Mamba achieves higher accuracy and, owing to its lower memory footprint, allows exploration of smaller patch sizes. This advantage is expected to become more significant for larger porous media (i.e., higher values of $n$) and for media with shorter spatial correlation lengths. According to Table \ref{TableMambaViT}, the ViT model generally requires less time per epoch, although the difference from the Vision Mamba model is not substantial.


At the end of this subsection, we provide a brief explanation of the ViT architecture implemented in this study. The network partitions each $64 \times 64 \times 64$ porous medium into non-overlapping patches (e.g., $8 \times 8 \times 8$ patches when the patch size is 8) and maps each patch to a token via a three-dimensional convolutional stem whose kernel and stride are equal to the patch size, producing embeddings of dimension 64. A learned absolute three-dimensional positional embedding, defined on a base token grid (e.g., $8 \times 8 \times 8$ grid for a patch size of 8), is trilinearly interpolated to the current token grid and added to the tokens. The encoder consists of three pre-normalized Transformer blocks; in each block, tokens are normalized, processed by multi-head self-attention with 8 heads, and merged back through a residual connection. This is followed by another normalization, a two-layer MLP with Gaussian error unit activation and dropout, and a second residual addition. After the block stack, tokens are layer-normalized and aggregated by global average pooling, and a linear head maps the pooled representation to a single scalar permeability prediction. Further details can be found in the original Transformer \cite{vaswani2023attentionneed} and Vision Transformer \cite{dosovitskiy2021imageworth16x16words} articles, as well as in our open-source code, the link to which is provided at the end of this article.

\begin{table}[htb]
 \centering
 \small
\caption{Performance comparison between Vision Mamba (ViM) and Vision Transformer (ViT). Reported metrics include the $R^2$ score, root mean square error (RMSE), minimum relative error (MiRE), and maximum relative error (MaRE) on the test set (170 samples) for different patch sizes. See text for details of the setup for each architecture. The symbol $\times$ indicates that the corresponding machine learning experiment could not be run due to GPU memory limitations.}

\label{TableMambaViT}
\begin{tabular}{l|ll|ll|ll|ll}
\toprule
Patch size & \multicolumn{2}{c}{4} & \multicolumn{2}{c}{8} & \multicolumn{2}{c}{16} & \multicolumn{2}{c}{32} \\

\midrule
 Model & ViM & ViT & ViM & ViT & ViM & ViT & ViM & ViT \\
\midrule
GPU memory  &  19.967 & $\times$ & 2.732 & 7.318 & 0.477 & 0.344 & 0.322 & 0.305 \\
per epoch (GB) &  &  &  & &  &  &  &  \\
\midrule
Training time  & 12.0 & $\times$ & 3.0 & 2.7 & 2.5 & 2.1 & 2.6 &  2.3  \\
per epoch (s) &  &  &  & &  &  &  &   \\
\midrule
Trainable  & 167169 & 187073 & 195841 & 215745 & 425217 & 445121 & 2260225 & 2280129 \\
parameters &  &  &  & &  &  &  &  \\
\midrule
$R^2$ score & 0.9934 & $\times$ & 0.9969 & 0.9838 & 0.9974 & 0.9957 & 0.9817 & 0.9491  \\
\midrule
RMSE (mD) & 3.9571 & $\times$ & 2.6939 & 6.1794 & 2.4557 & 3.1903 & 6.5718 & 10.9623  \\
\midrule
MiRE & 0.0001 & $\times$  & 0.0003 & 0.0001 & 0.0001 & 0.0001 & 0.0001 & 0.0001 \\
\midrule
MaRE & 0.4478 & $\times$  & 0.2708 & 0.6973 & 0.2105 & 0.3843 &  0.4586 & 0.7567 \\
\bottomrule
\end{tabular}
\end{table}


\begin{figure*}[!t]
  \centering

  \centerline{$N_{\text{block}}=1$ \hspace{5.5cm} $N_{\text{block}}=2$}
  \includegraphics[width=0.47\textwidth]{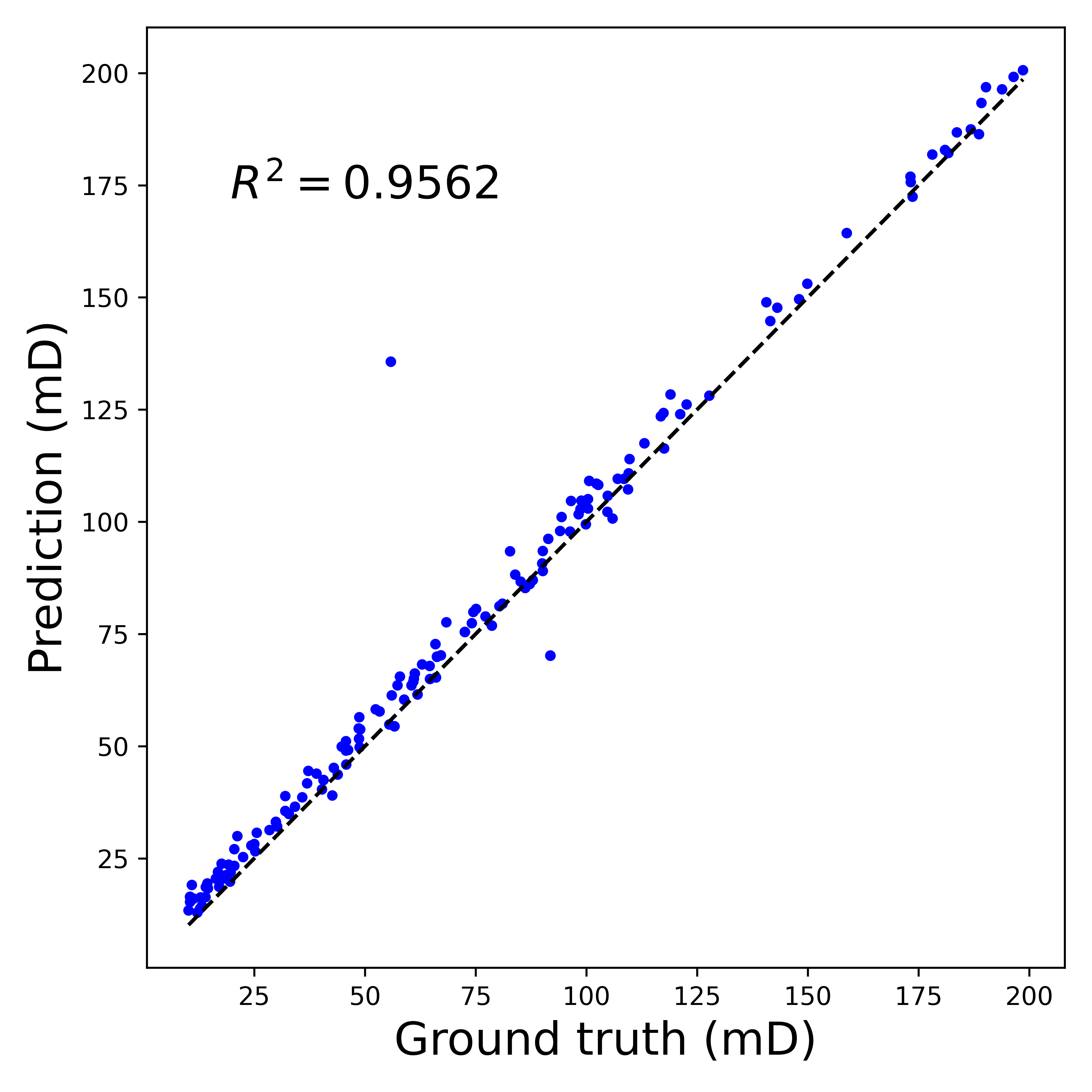}\hfill
  \includegraphics[width=0.47\textwidth]{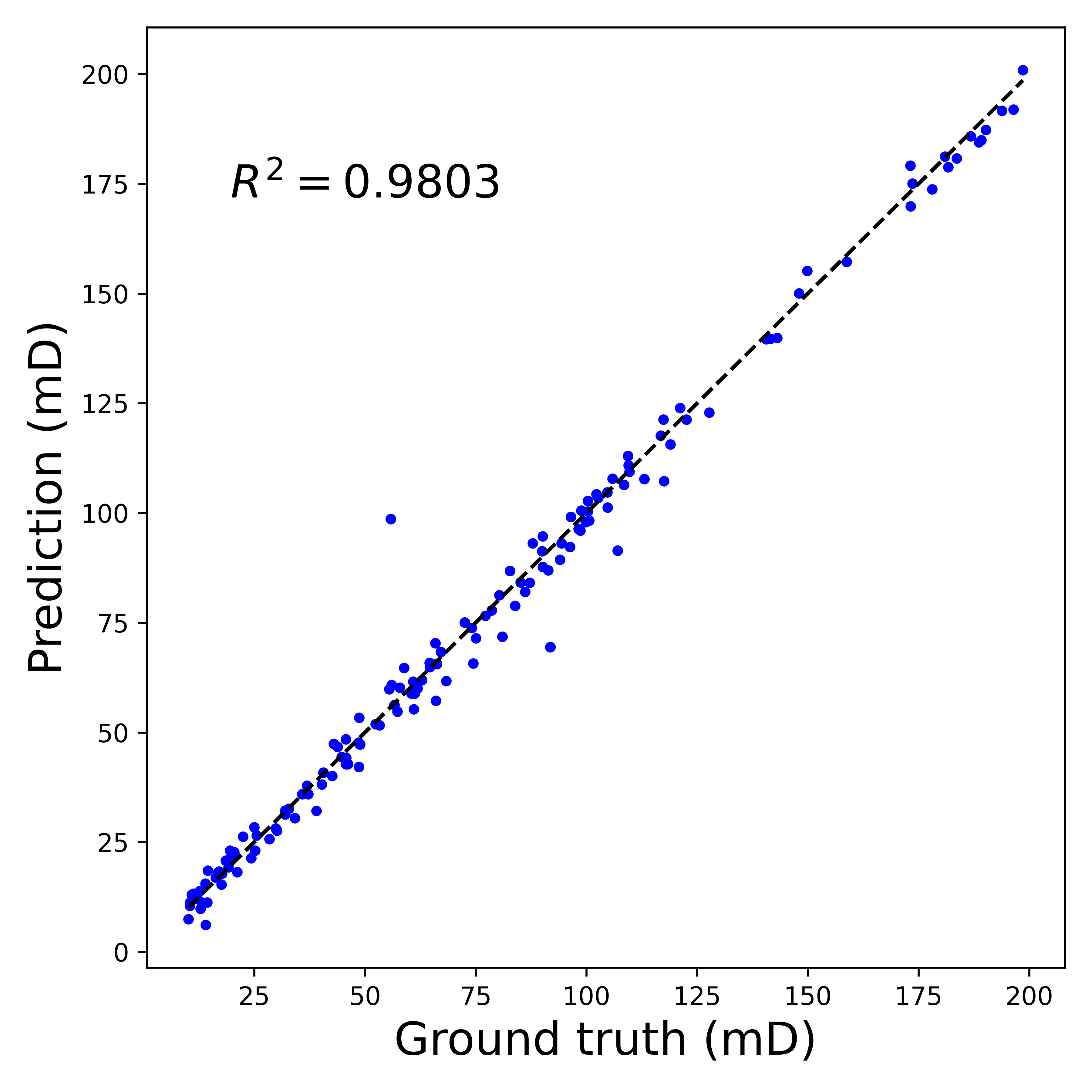}

\centerline{$N_{\text{block}}=4$ \hspace{5.5cm} $N_{\text{block}}=5$}
  \includegraphics[width=0.47\textwidth]{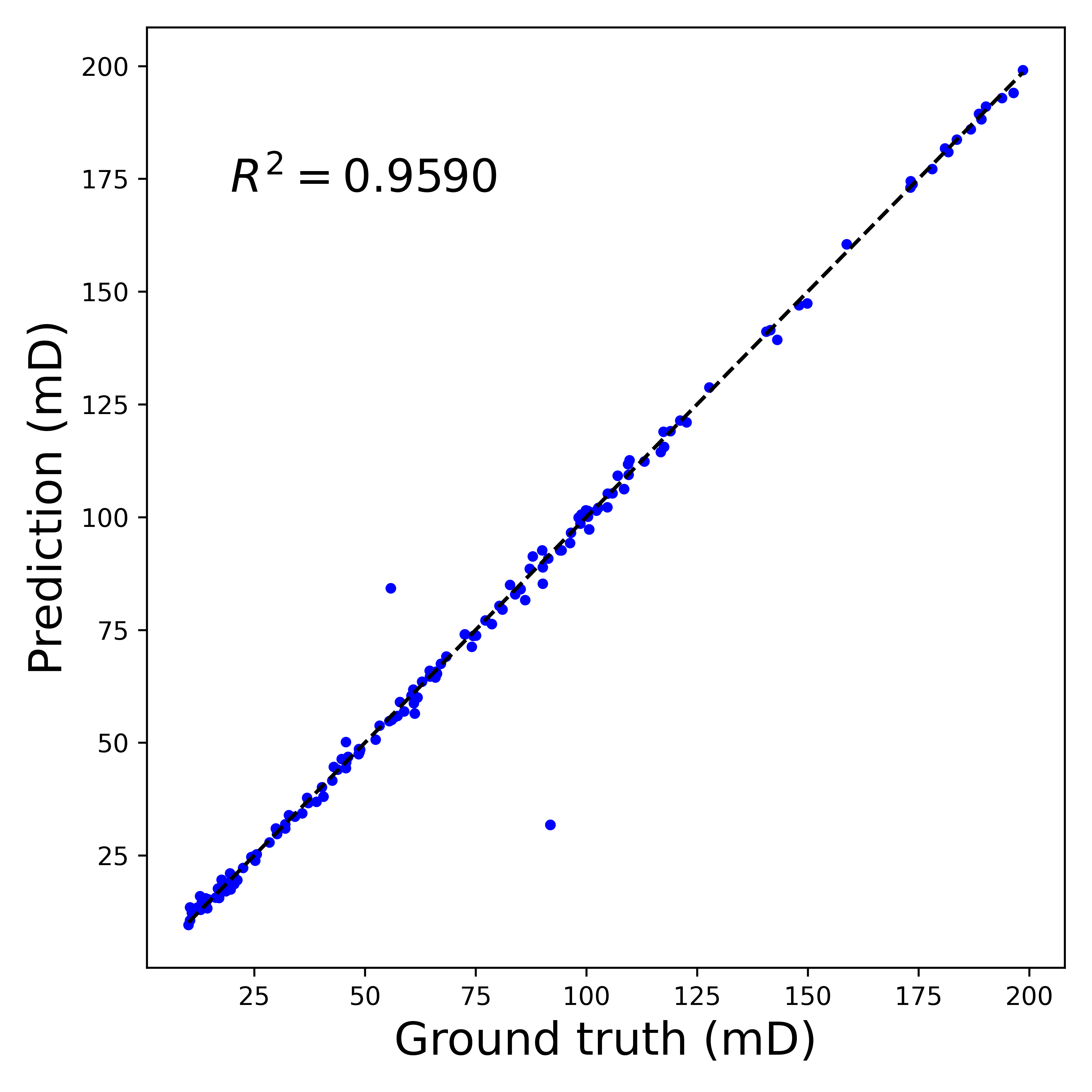}\hfill
  \includegraphics[width=0.47\textwidth]{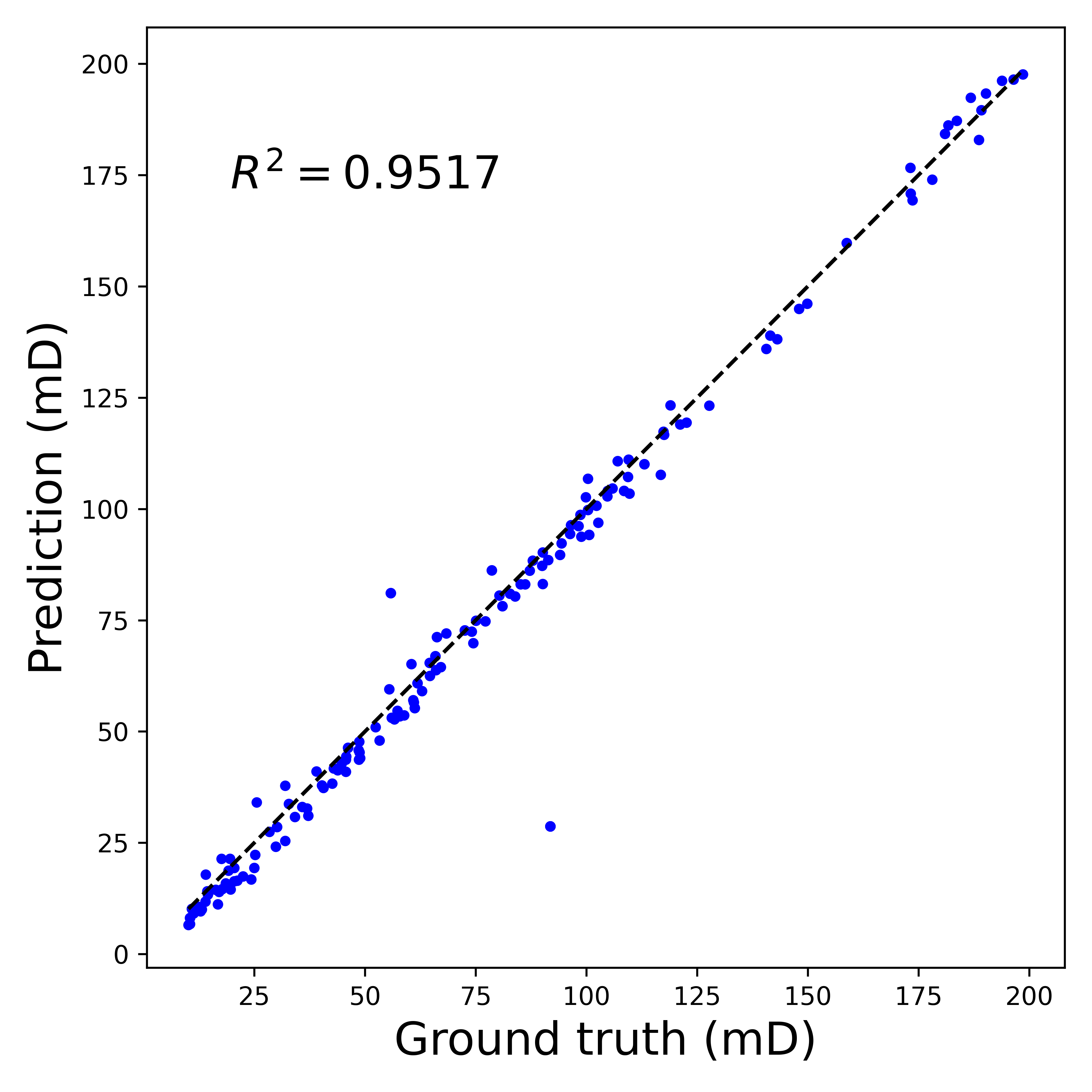}

  \caption{Performance of Vision Mamba for different numbers of blocks.}
  \label{Fig3}
\end{figure*}


\subsection{Ablation studies}
\label{Sect53}

In this section, our objective is to examine the influence of several key hyperparameters of the neural network on its performance in predicting the permeability of porous media. While designing a neural network, it is essential to perform hyperparameter fine-tuning to achieve the best possible performance, which is referred to as ablation studies. Interpreting the obtained results provides valuable insights into the behavior of the network for the current specific application in this article. For illustration, we focus on three important parameters. The first parameter is the number of Vision Mamba blocks ($N_{\text{block}}$). As explained in Sect. \ref{Sect4}, each block can be connected to the subsequent one, thereby progressively deepening the network. The results of this investigation are summarized in Table \ref{Table1}, where the number of Vision Mamba blocks ($N_{\text{block}}$) was varied from 1 to 5. For each configuration, we report the coefficient of determination (i.e., $R^2$ score), the root mean square error, as well as the maximum and minimum relative errors. As observed from Table \ref{Table1}, the best performance is obtained with a network consisting of three blocks (i.e., $N_{\text{block}}=3$), based on maximizing the $R^2$ score and minimizing the root mean square error on the test set (170 data). When the number of blocks is reduced, the network becomes shallower and the number of trainable parameters decreases, which leads to a decline in performance. However, this reduction is not particularly severe. For example, when the number of Vision Mamba blocks is reduced from 3 to 2, the $R^2$ score decreases only slightly from 0.9969 to 0.9803. Even with just a single Vision Mamba block (i.e., $N_{\text{block}}=1$), the $R^2$ score remains at 0.9562, indicating that the network maintains reasonable performance. Conversely, increasing the number of blocks from 3 to 4 and 5 leads to $R^2$ scores of 0.9590 and 0.9517, respectively. Thus, no performance improvement is observed beyond 3 Vision Mamba blocks; instead, a slight reduction in the $R^2$ score occurs. This decline may be attributed to a slight overfitting on the training data set, as the number of trainable parameters increases. This decline may be attributed to the increased capacity of the network and the number of trainable parameters, resulting in an unnecessarily large model for the current size of the training data.

\begin{table}[htb]
 \centering
 \small
\caption{$R^2$ score, root mean square error, and minimum/maximum relative errors of the test set (170 samples) for different numbers of Vision Mamba blocks in the proposed neural network. The patch size is fixed at 8.}\label{Table1}
\begin{tabular}{lllllll}
\toprule
 Number of Vision Mamba blocks ($N_{\text{block}}$) & 1 & 2 & 3 & 4 & 5  \\
\midrule
 $R^2$ score & 0.9562 & 0.9803 & 0.9969 & 0.9590 & 0.9517 \\
 Root mean square error (mD) & 10.1734 & 6.8171 & 2.6939 & 9.8449 & 10.6828 \\
 Minimum relative error & 0.0030 & 0.0001 & 0.0003 & 0.0001 & 0.0001 \\
 Maximum relative error & 1.4327 &  0.7685 & 0.2708 & 0.6538 & 0.6875\\
\bottomrule
\end{tabular}
\end{table}


The next hyperparameter we investigate is the patch size, the concept of which was explained in Sect. \ref{Sect3}. The outcomes of this investigation are listed in Table \ref{Table2}. Accordingly, we consider five different patch sizes: 4, 8, 16, 32, and 64. Similar to the previous case, we take the $R^2$ score and the root mean square error as benchmarks. Consequently, the best performance is obtained with a patch size of 16. However, the difference in $R^2$ scores across the different patch sizes is relatively small, with the highest being 0.9974 for the patch size of 16 and the lowest 0.9805 for the patch size of 64. 


As shown in Table \ref{Table2}, the overall performance of the network decreases slightly as the patch size increases beyond 16. This trend can be explained as follows. The patch size determines how each three-dimensional input is divided into smaller cubes and transformed into a sequence of patches, allowing the network to learn both their features and their relationships with neighboring patches. As discussed in Sect. \ref{Sect3}, these sequences are constructed by scanning the input along three spatial directions. In this sense, when the patch size is 64 and the porous media samples are also of size $64 \times 64 \times 64$ (i.e., $n=64$), no subdivision occurs; the entire cube is treated as a single patch, and fine-scale details are lost. In contrast, with a patch size of 8, a $64 \times 64 \times 64$ cube (i.e., $n=64$) is divided into 512 smaller patches, which are then sequentially processed in three spatial directions (e.g., length, width, and height). Hence, this representation enables the network to better capture local features, leading to more accurate permeability predictions in porous media. Additionally, we observe from Table \ref{Table2} that using the patch size of 8 does not improve performance and yields results nearly identical to those with a patch size of 16. It is conjectured that this behavior is related to the spatial correlation length of the dataset, which is 17 voxels (i.e., $\ell_c = 17$). This may suggest that the optimal patch size should be close to the spatial correlation length (if known), since it encapsulates the dominant information embedded at that scale.


\begin{table}[htb]
 \centering
 \small
\caption{$R^2$ score, root mean square error, and minimum/maximum relative errors of the test set (170 samples) for different patch sizes in the Vision Mamba model. The number of Vision Mamba blocks is fixed at three ($N_\text{block}=3$).}\label{Table2}
\begin{tabular}{llllll}
\toprule
Patch size & 4 & 8 & 16 & 32 & 64 \\
\midrule
$R^2$ score & 0.9934 &  0.9969 & 0.9974 & 0.9817 & 0.9805 \\
Root mean square error (mD) & 3.9571 & 2.6939 & 2.4557 & 6.5718 & 6.7914 \\
Minimum relative error & 0.0001 & 0.0003 & 0.0001 & 0.0001 & 0.0001\\
Maximum relative error & 0.4478 & 0.2708 & 0.2105 &  0.4586 & 0.8878\\
\bottomrule
\end{tabular}
\end{table}

As described in Sect. \ref{Sect3}, the proposed Vision Mamba–based network processes 3D porous-media cubes along the three spatial axes ($x$, $y$, and $z$) and aggregates the resulting features by averaging, as in Eq. \ref{eq:axis_fuse}. The network's output is the permeability in the $x$-direction (see Sect. \ref{Sect2} and Eq. \ref{perm}). To test whether scanning along the other two axes helps predict $x-$direction permeability, we conduct an ablation in which, instead of scanning along all three axes and averaging, we scan exclusively along a single axis ($x$ only, $y$ only, or $z$ only). The results in Table \ref{ScanDirection} show that, while $x$-only scanning is more accurate than $y$-only or $z$-only (as expected, given that the target is $x$-permeability), aggregating features from all three axes yields the best performance, with higher $R^2$ and lower root mean squared error. This outcome is consistent with the underlying physics, which indicates that the average velocity in Eq. \ref{perm} and the permeability in the $x$-direction depend on the full three-dimensional pore geometry. Therefore, solving the Stokes equations (see Eqs. \ref{Eq1}--\ref{Eq2}) in three dimensions is required even when estimating a directional permeability.

\begin{table}[htb]
 \centering
 \small
\caption{Comparison of the $R^2$ score, root mean square error, and minimum/maximum relative errors on the test set (170 samples) for different scan directions in Vision Mamba (see Eq. \ref{eq:axis_fuse}). We set $N_\text{block}=3$ and the patch size to 8.}\label{ScanDirection}
\begin{tabular}{lllll}
\toprule
Scan direction & All three axes &  $x-$axis & $y-$axis & $z-$axis \\
\midrule
$R^2$ score & 0.9969  & 0.9945 & 0.9829 & 0.9704 \\
Root mean square error (mD) & 2.6939 & 3.5980 & 6.3660 & 8.3650 \\
Minimum relative error & 0.0003 & 0.0001 & 0.0001 & 0.0001\\
Maximum relative error & 0.2708 & 0.3503 & 0.4187 & 0.9440 \\
\bottomrule
\end{tabular}
\end{table}


The final hyperparameter examined is the batch size ($\mathcal{B}$) during training. As shown in Table \ref{Table3}, a batch size of 128 ($\mathcal{B}=128$) yields the highest $R^2$ score and the lowest root mean square error when the patch size is fixed at 8 and the number of Vision Mamba blocks is set to 3. Larger batch sizes, such as 256, accelerate training but reduce performance, with the $R^2$ score dropping from 0.9969 to 0.9700, indicating decreased accuracy in predicting porous media permeability.


\begin{table}[htb]
 \centering
 \small
\caption{$R^2$ score, root mean square error, and minimum/maximum relative errors of the test set (170 samples) for different Batch sizes in the Vision Mamba model. The number of Vision Mamba blocks is fixed at three ($N_\text{block}=3$). The patch size is set to 8.}\label{Table3}
\begin{tabular}{llllll}
\toprule
Batch size ($\mathcal{B}$) & 4 & 16 & 32 & 128 & 256 \\
\midrule
$R^2$ score &  0.9750 & 0.9895 & 0.9911 & 0.9969 &  0.9700 \\
Root mean square error (mD) & 7.6883 & 4.9847 & 4.5980 & 2.6939 & 8.4248\\
Minimum relative error & 0.0004 & 0.0008 & 0.0001 & 0.0003 & 0.0001\\
Maximum relative error & 0.5114 & 0.6885 & 0.3712 & 0.2708 & 0.5845\\
\bottomrule
\end{tabular}
\end{table}




\section{Summary and future research projects}
\label{Sect6}

In this article, we presented a neural network based on Vision Mamba for predicting the permeability of three-dimensional porous media. We demonstrated the effectiveness of the proposed model using evaluation criteria such as the coefficient of determination, mean square error, and maximum and minimum relative errors. We discussed the advantages of Vision Mamba compared to CNNs and ViTs for permeability prediction. In particular, we showed that, relative to CNNs, Vision Mamba requires far fewer trainable parameters while achieving superior performance. Furthermore, we demonstrated that GPU memory usage in Vision Mamba scales linearly with patch size, whereas in ViTs it scales quadratically. As a result, under limited GPU memory, Vision Mamba was able to successfully execute the machine learning experiments with small patch sizes, whereas ViTs could not be trained due to insufficient memory when using the same small batch sizes. Finally, we explored the impact of key hyperparameters, including the number of Vision Mamba blocks, patch size within each block, and batch size, to highlight their influence on model performance.

In the present article, we used the classification branch of Vision Mamba, where the neural network input represents the geometry of the porous medium and the output is the permeability as a scalar value. As one idea for future projects, one could use the segmentation branch of Vision Mamba such that, although the input remains the same three-dimensional cube describing the porous-medium geometry, the output is the predicted velocity field within the pore space. Of course, after obtaining the velocity field, the permeability can also be computed. However, access to the full velocity field provides more information. This can be done in the form of fully supervised deep learning or in the form of weakly supervised deep learning, if only sparse observations from the velocity field are available, by enforcing the governing equations (e.g., Eqs. \ref{Eq1}--\ref{Eq2}) as a loss function for the Vision Mamba network.

Another promising research direction could be the development of large language and vision models for porous media based on Vision Mamba rather than transformer architectures. Such foundation models could handle variable-size and multimineral porous media, enabling the prediction of physical and geometrical features, and offering interactive environments to integrate images, codes, texts, and mathematical formulations within a unified framework.

\begin{backmatter}

\bmsection{Acknowledgment}
The authors of this research article gratefully acknowledge the sponsors of the Stanford Center for Earth Resources Forecasting (SCERF).

\bmsection{Data Availability}
\label{data}
The Python implementation is openly available in a public GitHub repository and can be accessed at \url{https://github.com/Ali-Stanford/Vision_Mamba_3D_Porous_Media}.


\end{backmatter}


\bibliography{sample}






\end{document}